\documentclass[final,3p,times,twocolumn]{elsarticle}

\usepackage{amssymb}

\usepackage{makecell}
\usepackage{amsmath}
\usepackage{float}
\usepackage[colorlinks=true]{hyperref}
\usepackage{multirow}
\usepackage{cleveref}
\usepackage[table]{xcolor}
\usepackage{arydshln}

\usepackage{pifont}
\newcommand{\cmark}{\ding{51}}%
\newcommand{\xmark}{\ding{55}}%

\journal{Computers \& Security}

\begin{document}

\begin{frontmatter}


\title{\textit{AttackNet}: Enhancing Biometric Security via Tailored Convolutional
Neural Network Architectures for Liveness Detection}

\author[macerata,karazin_compscience]{Oleksandr Kuznetsov}
\ead{kuznetsov@karazin.ua}

\author[karazin_math]{Dmytro Zakharov}
\ead{zamdmytro@gmail.com}

\author[macerata,marche]{Emanuele Frontoni}
\ead{emanuele.frontoni@unimc.it}

\author[marche]{Andrea Maranesi}
\ead{andrea.maranesi99@gmail.com}

\affiliation[macerata]{
            organization={Department of Political Sciences, Communication and International Relations, University of Macerata},
            addressline={Via Crescimbeni, 30/32},
            city={Macerata},
            postcode={62100},
            country={Italy}}
\affiliation[karazin_compscience]{
            organization={Department of Information and Communication Systems Security, School of Computer Sciences},
            addressline={4 Svobody Sq.},
            city={Kharkiv},
            postcode={61022},
            country={Ukraine}}
\affiliation[karazin_math]{
            organization={Department of Applied Mathematics, V.N. Karazin Kharkiv National University},
            addressline={4 Svobody Sq.},
            city={Kharkiv},
            postcode={61022},
            country={Ukraine}}
\affiliation[marche]{
            organization={Department of Information Engineering, Marche Polytechnic University},
            addressline={Via Brecce
Bianche 12},
            city={Ancona},
            postcode={60131},
            country={Italy}}


\begin{abstract}

Biometric security is the cornerstone of modern identity verification and
authentication systems, where the integrity and reliability of biometric samples is of paramount importance. This paper introduces \textit{AttackNet}, a bespoke Convolutional Neural Network architecture, meticulously designed to combat spoofing threats in biometric systems. Rooted in deep learning methodologies, this model offers a layered defense mechanism, seamlessly transitioning from low-level feature extraction to high-level pattern discernment. Three distinctive architectural phases form the crux of the model, each underpinned by judiciously chosen activation functions, normalization techniques, and dropout layers to ensure robustness and resilience against adversarial attacks. Benchmarking our model across diverse datasets affirms its prowess, showcasing superior performance metrics in comparison to contemporary models. Furthermore, a detailed comparative analysis accentuates the model's efficacy, drawing parallels with prevailing state-of-the-art methodologies. Through iterative refinement and an informed architectural strategy, \textit{AttackNet} underscores the potential of deep learning in safeguarding the future of biometric security.

\end{abstract}


\end{frontmatter}


\section{Introduction}
\label{section:introduction}

The digitization of numerous aspects of modern life has been accompanied by an increasing dependence on biometric systems for authentication, ranging from facial recognition at international borders to fingerprint scanning for smartphone access \citep{mobile_authentication,biometric_cryptosystems,biometric_authentication_cnn}. As a result, these systems' accuracy, speed, and, most importantly, security have become central pillars in the broader discussion about personal and public safety in the digital age \citep{security_ai}. While biometrics bring unparalleled convenience and personalized user experience, they also raise new challenges, notably, the threat of spoofing \citep{face_spoofing_attacks}. 

Spoofing attacks, where malicious actors attempt to deceive a biometric system using falsified data, have witnessed a considerable surge in sophistication \citep{handbook_biometric}. Facially, it has evolved from merely showing photographs to employing 3D masks and manipulated video recordings \citep{recent_progress}. Such advancements underscore a grave reality: if not fortified, contemporary biometric systems remain susceptible to breaches that could have severe real-world consequences \citep{robust_presentation_attack}.

In response to these concerns, the academic and industrial research communities have dedicated significant efforts to bolster the liveness detection capabilities of biometric systems \citep{overview_liveness_detection,survey_biometric_auth}. Liveness detection aims to ensure that the biometric data presented is from a living person and not a spoof artifact. The cornerstone of these studies has been the application of deep learning techniques, especially Convolutional Neural Networks (CNNs), providing unmatched accuracy and performance balance in image and pattern recognition tasks \citep{ieee_standard_liveness,mdn_liveness_detection,essential_mobile}.

In the constantly evolving area of biometric security, our research makes a multifaceted contribution, enhancing both the technological prowess and the practical applicability of liveness detection systems:

\begin{itemize}
\item \textbf{Development of a High-Performance System:} Our primary objective has been to develop a highly efficient system capable of operating seamlessly on less demanding computational platforms. This goal led to the creation of \textit{AttackNet}, a model characterized by low computational consumption while maintaining high accuracy. This feature is particularly critical for deployment in scenarios with limited hardware capabilities, such as mobile and embedded devices, making it a versatile tool against spoofing attacks.

\item \textbf{Comprehensive Evaluation:} Going beyond merely presenting a novel architecture, we embark on a rigorous evaluation journey. \textit{AttackNet} is benchmarked across multiple datasets and compared against traditional and state-of-the-art models. This exhaustive evaluation not only underscores the model's robustness but also offers insights into areas of potential improvement, laying the groundwork for future research in the domain.

\item \textbf{Cross-Database Liveness Detection:} Recognizing the evolving nature of threats and the diversity of environments in which biometric systems operate, we emphasize the importance of cross-database liveness detection. Our research delves into applying models trained on one dataset and tested on another, a rigorous test of the model's adaptability and generalizability.

\end{itemize}

By integrating these elements into our study, we aim to push the boundaries of biometric security research. Our goal is to advance the field technically and ensure that as technology becomes increasingly integrated into our daily lives, it remains a tool for empowerment, safeguarding against exploitation while being accessible and efficient even on less capable devices.

This paper is organized as follows: in \Cref{section:analysis}, we analyze the related work, positioning our research within the existing body of literature. \Cref{section:architecture} follows by a detailed description of the proposed \textit{AttackNet} architecture. Subsequent \Cref{section:methodology} presents the methodology of our experiments, the datasets employed, and our comprehensive evaluation results. In \Cref{section:discussion}, we enumerate the implications, strengths, and limitations of our findings. Finally, in \Cref{section:conclusion}, the paper concludes with key takeaways, potential future directions, and our closing remarks.

\section{Analysis of Related Works}\label{section:analysis}

Face anti-spoofing has been an area of high interest. Each study in this domain carries a unique perspective on various challenges, vulnerabilities, and potential mitigations. Nevertheless, the evolution of attack methodologies continues to open the gaps demanding attention. Here, we chronologically specify the works in this domain to gain insights into the state-of-the-art.

Sun et al. \citep{fcn_detection} pushed the boundaries of spoofing detection by the usage of depth-based Fully Convolutional Networks (FCNs). Their innovation, named the Spatial Aggregation of Pixel-level Local Classifiers (SAPLC), was a groundbreaking attempt to address vulnerabilities. However, their cross-database testing demonstrated a vulnerability with a Half Total Error Rate (HTER) nearing 30\% on the \textit{Replay Attack} dataset, pointing to a systemic challenge of model generalization across diverse datasets.

Alotaibi and Mahmood \citep{single_frame} ventured into deep convolutional neural networks, focusing on their efficacy in detecting spoofing assaults. By analyzing sequential frames and extracting pertinent features, they made notable strides, reflected in their 4\% HTER on the Replay Attack dataset. Yet, even with their sophisticated approach, the persistence of an HTER of this magnitude beckons for further enhancements in deep learning applications.

Chingovska et al. \citep{effectiveness_local_binary} delved into the intricacies of Local Binary Patterns (LBP),
examining their performance against various spoofing challenges. Their findings, which converged at an HTER of 15\%, encapsulated the moderate success of LBPs. This exploratory effort, while laudable, underscored the search for methodologies that could supersede LBPs
in efficacy.

Furthering the narrative of mask-based spoofing, Erdogmus and Marcel \citep{kinect} scrutinized the susceptibility of 2D face recognition systems against 3D facial masks. Their results, revealing a compelling 20\% HTER even with LBP-based countermeasures, highlighted the pressing requirement for fortified defenses against such sophisticated attacks.

The work of Bhattacharjee et al. \citep{csmad} offered a sobering perspective on the fragility of CNN-based face-recognition systems. Their exposition on vulnerabilities due to custom-made silicone masks presented a daunting picture. While their proposed solution gravitated towards a presentation attack detection method devoid of deep learning, it further emphasized the vast scope for integrating advanced machine learning techniques to combat such threats.

Building upon these findings, Chingovska et al. \citep{under_spoofing_attacks} comprehensively reported on the
diverse threat landscape, focusing on the vulnerabilities of face recognition systems operating across the visual spectrum. Their ventures into multispectral processing revealed a potential improvement, yet the persistence of HTER rates between 5-7\% signified the ongoing quest for optimization.

Kotwal and Marcel \citep{3d_nir}, in a refreshing turn, probed the capabilities of a pre-trained CNN equipped with a novel patch pooling mechanism. Targeting near-infrared imaging, their methodology displayed remarkable success against 3D mask attacks, particularly on the
WMCA dataset. Nevertheless, their focus's specificity on NIR and 3D masks underscores the vast realm of uncharted territories in other imaging modalities and attack vectors.

Mallat and Dugelay \citep{indirect_synthetic} introduced the community to a new frontier of attacks targeting thermal face recognition systems. Their revelations about the existing system vulnerabilities
painted a clear picture of the need to fortify defenses, particularly as adversaries continually innovate.

The work of \citep{visual_saliency} proposed a unique face anti-spoofing technique using visual saliency and facial motion characteristics against silicone mask attacks. Their accomplishments, outpacing many contemporary methods, still revealed a nearly 9\% HTER, indicating that the pursuit for the perfect countermeasure against such intricate attacks
remains ongoing. 

Concluding our review, Arora et al. \citep{framework_spoofing} steered the discourse towards feature extraction using pre-trained convolutional autoencoders. Their commendable HTER of 4\% showcased the potential of their methodology, yet the 40\% HTER in cross-database testing reaffirmed the overarching challenge of ensuring robust cross-dataset generalization.

In synthesis, the rich tapestry of face anti-spoofing research illustrates both our
community's achievements and the continual challenges that loom large. While each study has paved the way for subsequent advancements, the persistent vulnerabilities, inconsistencies in cross-dataset performances, and the emergence of novel, sophisticated attacks underline the imperative of continued innovation. It's within these gaps that our current research endeavors find their motivation and significance.

\section{The AttackNet Architecture: A Deep Dive}\label{section:architecture}

The increasing sophistication of spoofing attempts in biometric systems necessitates a specialized approach in model architecture. In response to this need, we introduce the \textit{AttackNet} architecture \citep{attacknet}, thoughtfully constructed to address the specific challenges of liveness detection. This section provides a comprehensive rationale for our architectural choices and discusses why certain advanced techniques from recent literature were not incorporated.

\subsection{Architectural Overview}

\textit{AttackNet} is designed with the core objectives of depth, feature integration, and regularization, each serving a distinct purpose:
\begin{enumerate}
\item \textbf{Depth for Complex Pattern Recognition:} The depth of the architecture, achieved through multiple convolutional layers, is crucial for extracting complex patterns that distinguish authentic biometric data from spoof attempts. Deep layers allow the model to learn hierarchies of features, which is fundamental in capturing the subtleties of biometric data.
\item \textbf{Feature Integration for Comprehensive Learning:} The model utilizes feature integration strategies, such as residual connections, to combine information across different layers. This approach ensures a comprehensive learning process, where both low-level and high-level features contribute to the final decision-making.
\item \textbf{Regularization for Robustness and Generalization:} Regularization techniques, including Batch Normalization and Dropout, are employed to prevent overfitting. This is particularly important to ensure that the model remains effective across varied datasets and real-world scenarios.
\end{enumerate}

\begin{figure*}
    \centering
    \includegraphics[width=0.7\textwidth]{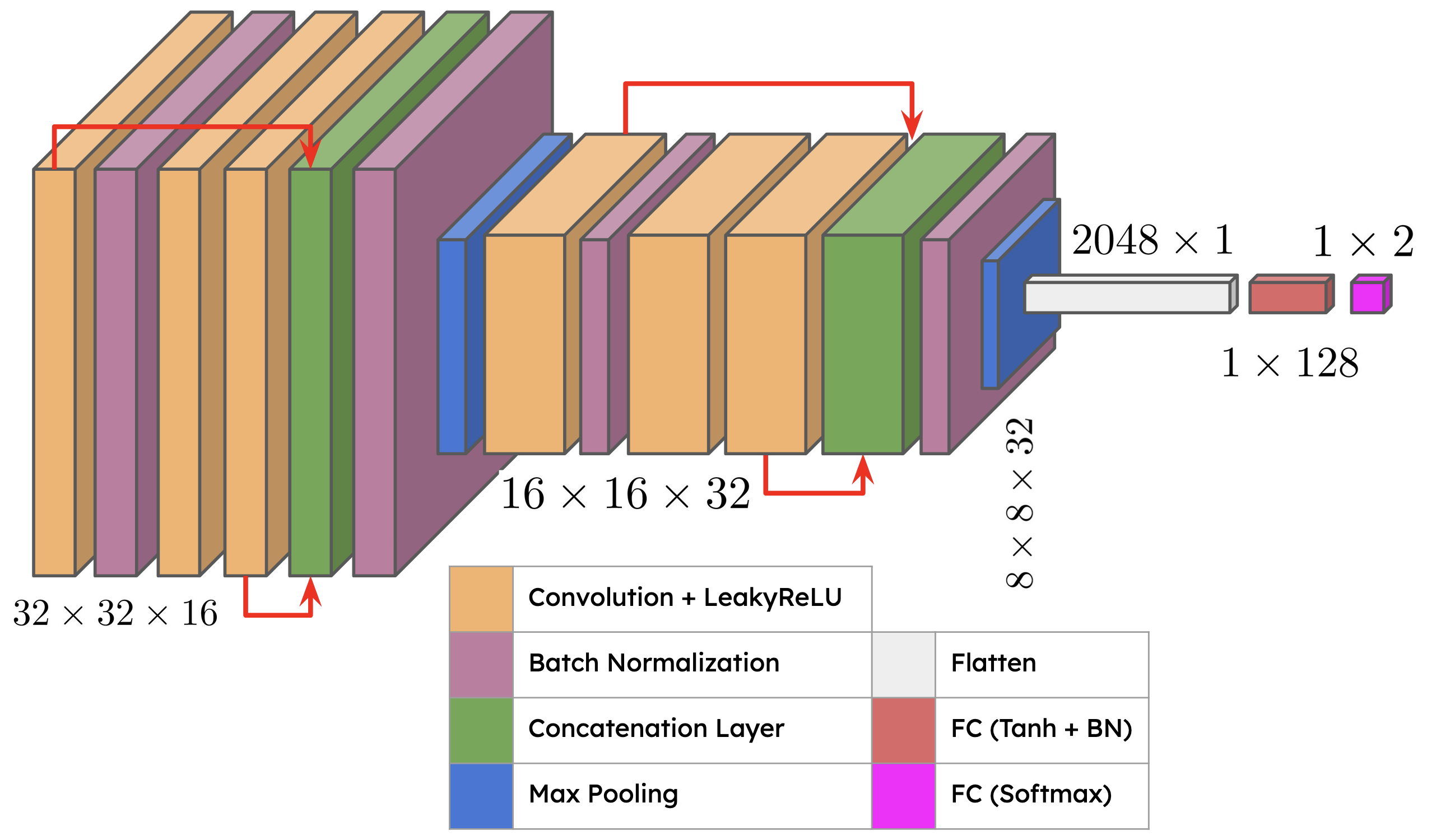}
    \caption{\textit{AttackNet} Architecture \citep{attacknet}}
    \label{fig:attacknet}
\end{figure*}

\Cref{fig:attacknet} provides a visual representation of the architecture, which we shall now dissect layer-by-layer.

\subsubsection{Initial Convolutional Phase}

The architecture commences with a convolutional phase composed of three convolutional layers, each employing a $3 \times 3$ kernel. The choice of a $3 \times 3$ kernel, a staple in advanced architectures like \textit{VGG}, allows for the gradual extraction of spatial hierarchies without losing critical information.

The model employs LeakyReLU as its activation function, which is defined as
\begin{equation}
    h(x;\alpha) \triangleq \max\{x,\alpha x\}, \; \alpha < 1,
\end{equation}

where $\alpha$ is a hyperparameter, typically significantly less than $1$. For $\alpha=0$ we get the traditional ReLU function. 

Unlike traditional ReLU, which can cause information loss for negative values, LeakyReLU allows a small gradient when the unit is not active, helping in preventing the dying ReLU problem. Indeed, if we consider the derivative of the LeakyReLU function, we obtain
\begin{equation}
    \frac{dh(x;\alpha)}{dx} = \begin{cases}
        1, & x > 0 \\
        \alpha, & x \leq 0
    \end{cases},
\end{equation}

from which we clearly see that for negative $x$, the gradient is non-zero and equals to $\alpha$.

Another noteworthy design choice is the use of residual connections. By adding the output from the initial layer to the third layer, we ensure that the network can learn identity functions for layers, which accelerates convergence and mitigates the vanishing gradient problem, especially vital given the depth of our network.

\subsubsection{Second Convolutional Phase}

Echoing the design considerations from the initial phase, the second convolutional phase similarly employs a sequence of convolutional layers with residual connections. This phase, however, doubles the number of filters, allowing the model to capture more intricate patterns as we go deeper into the network. Regularization, through Batch Normalization and Dropout, ensures that the model remains robust and avoids overfitting on the training data.

\subsubsection{Dense Phase}

The flattened output from the convolutional stack is then passed through a dense layer of $128$ neurons with $\tanh$ activation. This design choice ensures that the feature vectors' magnitude remains between $-1$ and $1$, providing a normalized and consistent feature representation before classification.

\subsubsection{Classification}

The final layer employs a softmax activation $\mathcal{S}: \mathbb{R}^C \to (0,1)^C$, which is defined as
\begin{equation}
\mathcal{S}(\mathbf{z})_c \triangleq \frac{\exp(z_c)}{\sum_{c'=1}^C \exp(z_{c'})}, \; c \in \{1,\dots,C\},
\end{equation}
to classify the input into one of the two categories - genuine or spoof. The binary classification aligns with the objective of liveness detection.

\subsection{Model Performance}

Due to the small number of parameters, the model exhibits a significant performance, making it a perfect choice for mobile and embedded devices. To compare our architecture with state-of-the-art models, we will use two parameters: \textit{MFLOP}s (million floating point operations) and a number of parameters, which are widely used as metrics for performance measurement. Table \ref{table:performance} illustrates the comparison with state-of-the-art models such as \textit{MobileNet} and \textit{ShuffleNet}. 

\begin{table*}
\centering
\begin{tabular}{c c c} 
 \Xhline{3\arrayrulewidth}
 \textbf{Model Architecture} & \textbf{MFLOPs} & \textbf{\# Parameters} \\ [0.5ex] 
 \textbf{AttackNet} & $\mathbf{22.7}$ & \textbf{0.3M}\\[0.5ex] \hdashline
 \textbf{MobileNetV3-Large 1.0} \citep{mobilenet_v3} & $109.5$ & 5.4M\\
 \textbf{MobileNetV3-Small 1.0} \citep{mobilenet_v3} & $28$ & 4.0M\\
 \textbf{MobileNetV2} \citep{mobilenet_v2} & $150$ & 3.4M
 \\[0.5ex]
 \hdashline
 \textbf{ShuffleNetV2 $1\times$} \citep{shufflenet_v2} & $146$ & 2.3M\\
 \textbf{ShuffleNetV2 $0.5\times$} \citep{shufflenet_v2} & $41$ & 1.4M\\
 \Xhline{3\arrayrulewidth}
\end{tabular}
\caption{Comparison of a model performance using MFLOPs and \# parameters with other state-of-the-art models}
\label{table:performance}
\end{table*}

The achieved efficiency of our \textit{AttackNet} model is noteworthy, as it registers at 22.7 \textit{MFLOP}s while only utilizing a modest 0.29k parameters. This efficiency level is particularly significant when contrasted with \textit{MobileNetV3-Small}, which necessitates a substantially higher number of parameters despite a comparable \textit{MFLOP} count. This disparity increases the overall weight of the \textit{MobileNetV3-Small} model. Such a factor is paramount in developing neural networks, especially for applications involving embedded devices where model size and computational efficiency are critical considerations.

To provide a more evident representation of the speed (although less universal), we have decided to launch the model \textbf{50000 times} on different images and record the inference time. After testing on \textit{MacBook M1}, we get that the average inferencing time is roughly \textbf{2.9 ms} (corresponding to approximately \textbf{345 FPS}).

In the contemporary landscape of technology, the significance of embedded devices, especially in biometric security and liveness detection, cannot be overstated. These devices are increasingly becoming integral in various applications, from personal smartphones to sophisticated security systems in the public and private sectors. Their limited computational resources demand models like \textit{AttackNet}, which are not only highly accurate but also lightweight and efficient.

Embedded systems are often at the frontline in the battle against spoofing attacks, being the primary interfaces in many biometric systems. The ability of these devices to process biometric data quickly and reliably hinges on the performance of the underlying models. In this context, the efficiency of \textit{AttackNet}, demonstrated through its low MFLOPs and minimal parameter count, is particularly pertinent.

Furthermore, the growing trend towards the Internet of Things (IoT) and smart devices underscores the need for such efficient models. As more devices connect and play a role in security and authentication, the demand for lightweight yet powerful AI models is set to increase exponentially. 

\subsection{Design Rationale}

The architectural choices in \textit{AttackNet} were guided by the following considerations:
\begin{itemize}
\item \textbf{Residual Connections:} These are instrumental in combating the vanishing gradient problem in deep networks, ensuring effective training and convergence.
\item \textbf{LeakyReLU Activation:} Selected over traditional ReLU to mitigate the dying ReLU issue, which can be a significant impediment in learning complex patterns.
\item \textbf{Batch Normalization and Dropout:} These regularization techniques are pivotal in ensuring that the model does not overfit to the training data, thereby enhancing its performance in diverse environments.
\end{itemize}

While recent literature has seen the emergence of sophisticated techniques like attention mechanisms and few-shot learning, their exclusion in \textit{AttackNet} was a considered decision. Our primary focus was on creating a model deeply tailored to liveness detection, where the primary challenge lies in recognizing subtle differences between authentic and spoofed biometric data. While techniques like attention mechanisms offer significant benefits in tasks requiring contextual understanding or focus, their utility in the specific context of liveness detection, as defined by our dataset and objectives, was not clear-cut. Our aim was to build a streamlined, efficient architecture with a clear focus on the fundamental task at hand, rather than incorporating additional complexities that may not contribute significantly to performance improvement in this specific application.

\subsection{Training Methodology}

The training of \textit{AttackNet} was structured to ensure optimal learning while avoiding overfitting. The model was trained using a supervised learning approach, with a dataset comprising labeled examples of both genuine and spoofed biometric data.

We employed a batch size of $64$, which ensured the stable gradient descent process and the model's ability to generalize well from the training data. The initial learning rate was set to $0.001$ together with the \textit{Adam optimizer}, helping in fine-tuning the model's weights and biases more accurately as it approaches the optimal solution. In the convolutional layers, we used $3 \times 3$ filters, which is generally considered effective in capturing the spatial hierarchies in the data without introducing excessive computational complexity. Dropout was employed at a rate of $0.5$ in the dense layers to prevent overfitting. Batch Normalization was also used after each convolutional layer to normalize the activations and speed up training.

The training was conducted over $100$ epochs, but we incorporated an early stopping mechanism based on the validation loss. Specifically, the training would terminate if the validation loss did not improve for ten consecutive epochs. This strategy helps in preventing overfitting by stopping the training process once the model ceases to learn further from the training data.

To enhance the model's ability to generalize, we applied data techniques, including random rotations, width and height shifts, shear transformations, and zoom augmentation. These techniques help in creating a robust model that is less sensitive to variations in input data.

Thus, the design of \textit{AttackNet} is a thoughtful response to the unique challenges posed by liveness detection in biometric systems. Each component of the architecture is carefully chosen to contribute towards a model that is both deep and intricate in its learning capabilities, yet robust and generalizable across diverse scenarios. The subsequent sections will detail the rigorous testing of \textit{AttackNet}, highlighting its performance and adaptability in comparison to other contemporary architectures.

\section{Methodology}\label{section:methodology}

To provide a comprehensive evaluation of the \textit{AttackNet} architecture and to position it within the wider landscape of liveness detection solutions, a systematic methodology was devised. Our approach aimed to probe not just the raw performance metrics but also the model's adaptability, resilience to overfitting, and behavior across a variety of scenarios.

\subsection{Datasets}

\begin{figure*}
    \centering
    \includegraphics[width=0.65\textwidth]{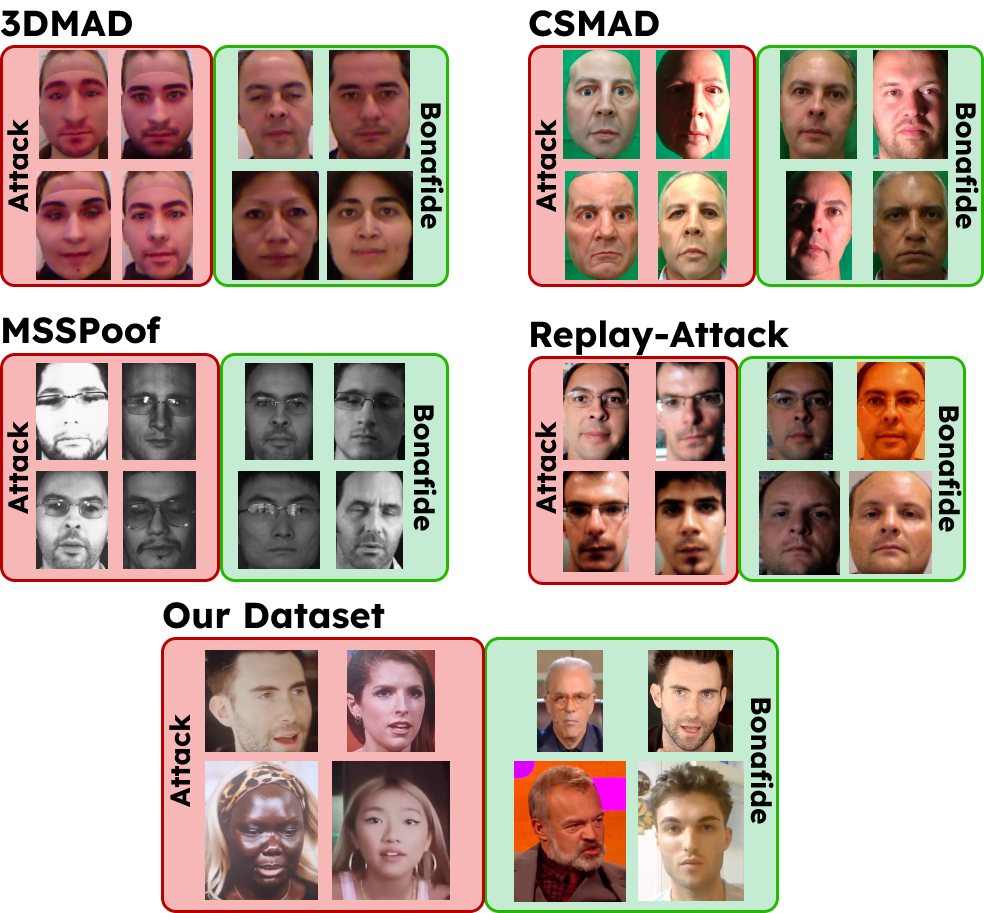}
    \caption{Datasets employed: \citep{csmad_dataset,kinect,under_spoofing_attacks,effectiveness_local_binary} and ours. In \textcolor{red}{red} we marked example attack images, and in \textcolor{green}{green} -- bonafide ones.}
    \label{fig:dataset}
\end{figure*}

Five primary datasets were utilized to facilitate a holistic assessment of the \textit{AttackNet} architecture and its efficiency across diverse attack vectors. We illustrate example pictures from all these datasets in \Cref{fig:dataset}.

\subsubsection{The Custom Silicone Mask Attack Dataset (CSMAD)}

Originating from the \textit{Idiap Research Institute}, the CSMAD is an invaluable resource for face presentation attack detection \citep{csmad,csmad_dataset}. It boasts presentation attacks leveraging six distinct, custom-made silicone masks, each costing approximately \textit{4000 USD}. This dataset includes \textbf{14 unique subjects}, designated by alphabetic codes ranging from $A$ to $N$, with subjects $A$ through $F$ also serving as targets. Their face data was vital in constructing their corresponding silicone masks. Besides, the dataset incorporates \textbf{diverse lighting conditions to augment variability}: fluorescent ceiling light and halogen lamps (both from the left and right individually and together), against a consistent green backdrop. 

In terms of the amount of material, \textit{CSMAD} includes \textbf{$87$ videos + $17$ JPG stills} of ``Bonafide'' class captured via a \textit{Nikon Coolpix} digital camera and \textbf{$159$ videos} of ``Attack'' class, further categorized into ``WEAR'' ($108$ videos where a mask is worn) and ``STAND'' ($51$ videos with masks mounted on a stand).

\textit{CSMAD}'s meticulous organization also includes a 'protocols' directory detailing the experimental procedures, vital for replicating the described vulnerability analysis of face-recognition systems.

\subsubsection{The 3D Mask Attack Database (3DMAD)}

The 3DMAD is a biometric spoofing treasure trove \citep{kinect}. Packed with $76500$ frames across $17$ individuals, it was captured using the \textit{Kinect} technology to record genuine and spoofed presentations. Each frame's components are an $11$-bit depth image ($640\times 480$ pixels), an $8$-bit RGB image ($640\times 480$ pixels), and manually annotated eye positions pertinent to the RGB image. 

The dataset includes structured recordings spanning three sessions, with two earmarked for authentic accesses and a third specifically for capturing 3D mask attacks. It also has precise eye-position labels for select frames, ensuring accuracy and consistency, together with real-size masks sourced from \textit{``ThatsMyFace.com''} alongside the original images used for their creation.

\subsubsection{The Multispectral-Spoof Face Spoofing Database}

\textit{MSSpoof} is yet another contribution from the \textit{Idiap Research Institute}, which is pivotal for assessing spoofing attacks \citep{under_spoofing_attacks}. Captured with a high-resolution \textit{uEye} camera, it boasts both \textit{VIS} and \textit{NIR} images under different lighting conditions. Detailed features comprise \textbf{$21$ distinct clients}, with recordings of \textbf{$70$ actual accesses} each, spanning $35$ VIS and $35$ NIR images across varied environments, and \textbf{$144$ spoofing attacks} per client. It is also split into train ($9$ clients), development ($6$ clients), and test ($6$ clients) sets to ensure unbiased evaluations. Moreover, it has manual annotations for every sample, pinpointing \textbf{$16$ key facial landmarks}.

\subsubsection{The Replay-Attack Database}

The \textit{Replay-Attack Database} is meticulously designed to address the challenges posed by face spoofing, comprising a significant \textbf{$1300$ video clips}, which are segmented into photo and video attack attempts from a cohort of \textbf{$50$ distinct clients} \citep{effectiveness_local_binary}. These have been recorded under diverse illumination environments. This compilation is a work by the \textit{Idiap Research Institute, Switzerland}.

To ensure the integrity of the training process, any client appearing in one segment is intentionally excluded from the others. The video captures two scenarios: authentic clients accessing a laptop through its built-in webcam and the portrayal of a photograph or video clip of the same client for a minimum duration of $9$ seconds. These videos, boasting a resolution of $320\times 240$ pixels, were captured on a \textit{Macbook} laptop employing the \textit{QuickTime} framework. While the video utility is native to \textit{Apple} computers, compatibility is ensured across multiple platforms, including \textit{Linux} and \textit{MS Windows}, through utilities such as \textit{mplayer} and \textit{ffmpeg}.

To enrich the database, real client access and the data garnered for attacks were recorded under two lighting conditions: (a) \textit{controlled} -- a consistent setting with office lights on, blinds drawn, and a uniform background, and (b) \textit{dynamic setup} -- blinds open, a more intricate background and office lights turned off.

Various attacking modes have been cataloged in the database. These include attacks using devices like \textit{iPhone 3GS}, \textit{iPad} (first generation), and printed hard-copy assaults. The comprehensive distribution of real accesses and attacks is detailed meticulously across the training, development, test, and enrollment sets. Additionally, this dataset is complemented with automatic face location annotations based on \textit{Modified Census Transform} (MCT).

\subsection{Our Proprietary Dataset}

Our proprietary dataset is a testament to the evolutionary advancements in antispoofing research \citep{attacknet}. It is split into two image sets -- bonafide and attackers. The bonafide set encapsulates images from video recordings of authentic individuals captured through smartphones or downloaded from the vast expanse of the internet. Conversely, the attacker set is populated with images from videos recorded through laptop webcams, wherein these laptops showcase the playback of the bonafide video set. The architecture of our dataset is outlined in \Cref{table:our_dataset}.

\begin{table}[H]
\begin{center}
\begin{tabular}{c c} 
 \Xhline{3\arrayrulewidth}
 \textbf{Attribute} & \textbf{Details} \\
 Dataset Dimensions & $4656$\\ 
 Class Distribution & $50/50$\\ 
 Training/Validation Split & $48/52$\\ 
Training Images & $2238$\\ 
Validation Images & $2418$\\ 
Training Class Distribution & $50/50$\\
Validation Class Distribution & $50/50$\\ \Xhline{3\arrayrulewidth}
\end{tabular}
\caption{
    \label{table:our_dataset}
    Our Dataset Structure
}
\end{center}
\end{table}

The core of our dataset is a rich assortment of videos primarily sourced from \textit{YouTube}, spotlighting genuine individuals. Subsequently, the same videos were recorded using a mobile device and broadcasted on auxiliary devices, such as laptops or other mobile device. This structure, featuring \textbf{$84$ videos}, is stratified into training and testing segments, with a balanced representation from both bonafide and attacker categories. This balance was achieved through meticulous under-sampling of frames from each video.

Collectively, these datasets offer an unprecedented breadth and depth of data, allowing for rigorous evaluations and facilitating advancements in biometric security.

\subsection{Performance Metrics}

When evaluating the performance of biometric systems, especially in face spoofing, a comprehensive understanding of the various metrics is essential. This ensures that the system's capabilities and potential vulnerabilities are correctly assessed. Below, we detail the metrics employed in our study.

\subsubsection{Confusion Matrix}

Before delving into the metrics, it is pivotal to understand the confusion matrix, a tool often used in the classification field to visualize the performance of an algorithm. It consists of (1) True Positives (TP) -- correctly identified genuine attempts, (2) True Negatives (TN) -- correctly identified attack attempts, (3) False Positives (FP) -- incorrectly identified genuine attempts as attacks, (4) False Negatives (FN) -- incorrectly identified attack attempts as genuine. See \Cref{table:Confusion_Matrix} for illustration.

\begin{table}[H]
\begin{center}
\begin{tabular}{c|c|c|} 
 \cline{2-3}
 & \multicolumn{2}{c|}{\cellcolor{gray!25}\textbf{Predicted}}  \\ \cline{2-3}
 & \cellcolor{green!25}\textbf{Genuine} & \cellcolor{red!25}\textbf{Attack} \\ [0.5ex] 
 \cline{2-3}\hline
 \multicolumn{1}{|c|}{\cellcolor{green!25}\textbf{Actual Genuine}} & TP & FN \\ \hline
 \multicolumn{1}{|c|}{\cellcolor{red!25}\textbf{Actual Attack}} & FP & TN \\ \hline
\end{tabular}
\caption{
    \label{table:Confusion_Matrix}
    Confusion Matrix
}
\end{center}
\end{table}

The confusion matrix is an essential tool in classification tasks, offering a detailed breakdown of a model's performance by accurately classifying instances into true positives, true negatives, false positives, and false negatives. This clear categorization helps in identifying the model's strengths and weaknesses. Notably, it reveals the symmetry of hypotheses in classification: swapping the roles of the classes (e.g., Genuine and Attack) in biometric systems would result in the interchange of key metrics like the False Acceptance Rate (FAR) and the False Rejection Rate (FRR). Such symmetry emphasizes the need for a balanced model evaluation. Additionally, combining the confusion matrix with metrics like Precision, Recall, and the Half Total Error Rate (HTER) provides a comprehensive framework for assessing the system's effectiveness, ensuring it reliably detects genuine and fraudulent instances.

\subsubsection{Precision}

Precision captures the ratio of correctly predicted positive observations to the total predicted positives:
\begin{equation}
\text{Precision} \triangleq \frac{\text{TP}}{\text{TP} + \text{FP}}.
\end{equation}

High precision relates to a low false positive rate. In security contexts, high precision indicates that it will likely be correct when a system grants access. However, it doesn't capture the system's robustness against attacks, which is critical in biometrics.

\subsubsection{Recall (Sensitivity)}

Recall calculates the ratio of correctly predicted positive observations to all the observations in the actual class:
\begin{equation}
    \text{Recall} \triangleq \frac{\text{TP}}{\text{TP}+\text{FN}}.
\end{equation}

It represents the system's ability to identify genuine attempts correctly. While a crucial metric, a system with high recall but low precision can be problematic. Such a system would catch most genuine attempts but might also grant access to many attackers.

\subsubsection{F1 Score}

The $F_1$ Score is the harmonic mean of $\text{Precision}$ and $\text{Recall}$ and provides a balanced view between the two metrics:
\begin{equation}
F_1 \triangleq \frac{2 \times \text{Precision} \times \text{Recall}}{\text{Precision} + \text{Recall}}
\end{equation}
It's handy when class distribution is uneven. When dealing with datasets where one type of classification is more critical than the other (e.g., preventing false negatives might be more crucial than preventing false positives), the $F_1$ score provides a more balanced metric.

\subsubsection{False Acceptance Rate (FAR) and False Rejection Rate (FRR)}

$\text{FAR}$ quantifies the likelihood that the system erroneously grants access to an attacker:
\begin{equation}
\text{FAR} \triangleq \frac{\text{FP}}{\text{FP}+\text{TN}}
\end{equation}

$\text{FRR}$ indicates the probability that the system wrongly denies access to a genuine user:
\begin{equation}
\text{FRR} \triangleq \frac{\text{FN}}{\text{TP}+\text{FN}}
\end{equation}

These are pivotal in biometric systems. A high $\text{FAR}$ might mean that attackers can breach the system easily, while a high $\text{FRR}$ might inconvenience genuine users. A delicate balance between the two is paramount.

\subsubsection{HTER (Half Total Error Rate)}

$\text{HTER}$ offers an overall system error rate by averaging $\text{FAR}$ and $\text{FRR}$:
\begin{equation}
    \text{HTER} \triangleq \frac{\text{FAR}+\text{FRR}}{2}
\end{equation}

This metric is beneficial in providing a singular, comprehensive measure of system accuracy.

It's an excellent overall metric for assessing the trade-offs between $\text{FAR}$ and $\text{FRR}$. In applications where both genuine user access and security are equally important, $\text{HTER}$ provides a singular measure to gauge system performance.

In conclusion, while each metric has significance, understanding the application context and associated risks is crucial. No single metric can evaluate a system's performance holistically, making a comprehensive analysis employing multiple metrics essential.

\section{Research Results}\label{section:results}

We conducted extensive cross-database testing to evaluate the robustness and adaptability of our proposed model, \textit{AttackNet}. This process involved training the model on one dataset and assessing its performance on different datasets. This approach is crucial for understanding how well a model generalizes to new, unseen data, which is a critical aspect of real-world application scenarios. The results of this rigorous testing are summarized in Table \ref{table:Cross_Database}, presenting key performance metrics such as Precision, Recall, $F_1$ Score, FAR, FRR, and HTER.

\subsection{Cross-Database Testing Results}

The cross-database testing results provide insightful revelations about the model's performance dynamics. We present a more nuanced discussion of these results, underlining the significance of certain numerical indicators.

\begin{itemize}
\item \textbf{MSSpoof Dataset:} When trained on the MSSpoof dataset, the model achieved high precision (0.96 for Bonafide and 0.93 for Attacker) and recall (0.92 for Bonafide and 0.96 for Attacker) on the same dataset. However, a notable dip in performance was observed on other datasets, with precision dropping to 0.56 for Bonafide attempts on the 3DMAD dataset. The most significant concern was the high FAR (up to 0.71 on Replay Attack Dataset), indicating a pronounced vulnerability in accurately distinguishing spoof attempts.

\item \textbf{3DMAD Dataset:} The model, when trained on the 3DMAD dataset, showed exceptional results (Precision and Recall of 1.0 for both Bonafide and Attacker) on the same dataset. Nevertheless, its performance varied when applied to other datasets, with a noticeable decrease in $F_1$ score (0.45 for Bonafide and 0.69 for Attacker on MSSpoof), highlighting challenges in maintaining accuracy across different data types.

\item \textbf{CSMAD Dataset:} Training on the CSMAD dataset resulted in reasonable effectiveness across datasets. However, variations in performance metrics, such as an FRR of 0.64 for Bonafide on MSSpoof, pointed toward inconsistencies in the model's predictive capabilities.

\item \textbf{Attack dataset:} Training on the Replay Attack dataset led to a high HTER of 0.48 when tested on MSSpoof but an impressively low HTER of 0.045 on its own dataset. This disparity underscores the model's potential in familiar environments but also its limitations in new contexts.

\item \textbf{Our Dataset:} The results from training on Our Dataset were mixed. While the model showed moderate generalization (HTER of 0.305 on Replay Attack), the elevated FAR (up to 0.87 on MSSpoof) and FRR (up to 0.87 on 3DMAD) in some instances indicated a need for further model refinement.

\end{itemize}

These findings emphasize the critical need for robust and versatile model designs capable of handling real-world data's diverse and complex nature. The numerical indicators, primarily the elevated error rates (FAR and FRR), highlight areas where the model requires optimization to improve its generalization abilities.

The critical findings of our research cast a spotlight on a fundamental challenge faced by AI models in the realm of spoofing attack detection. While a model may exhibit commendable performance on the dataset it was trained on, its efficacy noticeably dwindles when confronted with unfamiliar datasets. This phenomenon, which was observed consistently across our testing scenarios, forms the cornerstone of our study and provides vital insights into the limitations of AI models in varied spoofing scenarios.

Our future research will focus on addressing this limitation, exploring methodologies and techniques that can enhance the generalization capabilities of AI models in the context of liveness detection. The goal is to develop models that excel in familiar environments and demonstrate resilience and adaptability in the face of new and varied spoofing scenarios. This endeavor is crucial for advancing the reliability and effectiveness of AI systems in thwarting spoofing attacks across diverse real-world applications.

\begin{table*}
  \centering
  \begin{tabular}{||c|c|c|c|c|c|c|c|c|c|c|c||}
  \Xhline{3\arrayrulewidth}
     \multirow{8}{*}{\cellcolor{blue!25}{\rotatebox[origin=c]{90}{\textbf{MSSPoof}}}} & \multirow{2}{*}{\textbf{Metric}} & \multicolumn{2}{c|}{\cellcolor{blue!25}{\textbf{MSSpoof}}} & \multicolumn{2}{c|}{\cellcolor{green!25}{\textbf{3DMAD}}} & \multicolumn{2}{c|}{\cellcolor{yellow!25}{\textbf{CSMAD}}} & \multicolumn{2}{c|}{\cellcolor{red!25}{\textbf{Our Dataset}}} & \multicolumn{2}{c||}{\cellcolor{gray!25}{\textbf{Replay Attack}}} \\ \cline{3-12}
    &  & \textit{B} & \textit{A} & \textit{B} & \textit{A} & \textit{B} & \textit{A} & \textit{B} & \textit{A} & \textit{B} & \textit{A} \\ \cline{2-12}
    & Precision & $0.96$ & $0.93$ & $0.56$ & $0.68$ & $1.00$ & $0.65$ & $0.59$ & $0.58$ & $0.53$ & $0.51$ \\\cline{2-12}
    & Recall & $0.92$ & $0.96$ & $0.83$ & $0.36$ & $0.47$ & $1.0$ & $0.54$ & $0.64$ & $0.29$ & $0.74$ \\\cline{2-12}
    & $F_1$ Score & $0.94$ & $0.94$ & $0.67$ & $0.47$ & $0.64$ & $0.79$ & $0.56$ & $0.60$ & $0.37$ & $0.60$ \\\cline{2-12}
    & FAR & $0.08$ & $0.04$ & $0.44$ & $0.17$ & $0.53$ & $0.00$ & $0.46$ & $0.37$ & $0.71$ & $0.26$ \\\cline{2-12}
    & FRR & $0.04$ & $0.08$ & $0.17$ & $0.44$ & $0.00$ & $0.53$ & $0.37$ & $0.46$ & $0.26$ & $0.71$ \\\cline{2-12}
    & HTER & \multicolumn{2}{c|}{$0.060$} & \multicolumn{2}{c|}{$0.305$} & \multicolumn{2}{c|}{$0.265$} & \multicolumn{2}{c|}{$0.415$} & \multicolumn{2}{c||}{$0.485$} \\\hline\hline
    \multirow{8}{*}{\cellcolor{green!25}{\rotatebox[origin=c]{90}{\textbf{3DMAD}}}} & \multirow{2}{*}{\textbf{Metric}} & \multicolumn{2}{c|}{\cellcolor{blue!25}{\textbf{MSSpoof}}} & \multicolumn{2}{c|}{\cellcolor{green!25}{\textbf{3DMAD}}} & \multicolumn{2}{c|}{\cellcolor{yellow!25}{\textbf{CSMAD}}} & \multicolumn{2}{c|}{\cellcolor{red!25}{\textbf{Our Dataset}}} & \multicolumn{2}{c||}{\cellcolor{gray!25}{\textbf{Replay Attack}}} \\ \cline{3-12}
    &  & \textit{B} & \textit{A} & \textit{B} & \textit{A} & \textit{B} & \textit{A} & \textit{B} & \textit{A} & \textit{B} & \textit{A} \\ \cline{2-12}
    & Precision & $0.74$ & $0.57$ & $1.00$ & $1.00$ & $1.00$ & $0.79$ & $0.82$ & $0.63$ & $0.82$ & $0.63$ \\\cline{2-12}
    & Recall & $0.32$ & $0.89$ & $1.00$ & $1.00$ & $0.73$ & $1.00$ & $0.47$ & $0.90$ & $0.47$ & $0.90$ \\\cline{2-12}
    & $F_1$ Score & $0.45$ & $0.69$ & $1.00$ & $1.00$ & $0.84$ & $0.88$ & $0.60$ & $0.74$ & $0.60$ & $0.74$ \\\cline{2-12}
    & FAR & $0.68$ & $0.11$ & $0.00$ & $0.00$ & $0.27$ & $0.00$ & $0.53$ & $0.10$ & $0.53$ & $0.10$ \\\cline{2-12}
    & FRR & $0.11$ & $0.68$ & $0.00$ & $0.00$ & $0.00$ & $0.27$ & $0.10$ & $0.53$ & $0.10$ & $0.53$ \\\cline{2-12}
    & HTER & \multicolumn{2}{c|}{$0.395$} & \multicolumn{2}{c|}{$0.000$} & \multicolumn{2}{c|}{$0.135$} & \multicolumn{2}{c|}{$0.315$} & \multicolumn{2}{c||}{$0.315$} \\\hline\hline
    \multirow{8}{*}{\cellcolor{yellow!25}{\rotatebox[origin=c]{90}{\textbf{CSMAD}}}} & \multirow{2}{*}{\textbf{Metric}} & \multicolumn{2}{c|}{\cellcolor{blue!25}{\textbf{MSSpoof}}} & \multicolumn{2}{c|}{\cellcolor{green!25}{\textbf{3DMAD}}} & \multicolumn{2}{c|}{\cellcolor{yellow!25}{\textbf{CSMAD}}} & \multicolumn{2}{c|}{\cellcolor{red!25}{\textbf{Our Dataset}}} & \multicolumn{2}{c||}{\cellcolor{gray!25}{\textbf{Replay Attack}}} \\ \cline{3-12}
    &  & \textit{B} & \textit{A} & \textit{B} & \textit{A} & \textit{B} & \textit{A} & \textit{B} & \textit{A} & \textit{B} & \textit{A} \\ \cline{2-12}
    & Precision& $0.57$ & $0.53$ & $0.56$ & $1.0$ & $0.79$ & $1.0$ & $0.5$ & $0.49$ & $0.51$ & $0.51$ \\\cline{2-12}
    & Recall & $0.36$ & $0.72$ & $1.0$ & $0.21$ & $1.0$ & $0.73$ & $0.68$ & $0.31$ & $0.57$ & $0.45$\\\cline{2-12}
    & $F_1$ Score & $0.44$ & $0.61$ & $0.72$ & $0.35$ & $0.88$ & $0.84$ & $0.58$ & $0.38$ & $0.54$ & $0.48$  \\\cline{2-12}
    & FAR & $0.64$ & $0.28$ & $0.00$ & $0.79$ & $0.00$ & $0.27$ & $0.32$ & $0.69$ & $0.43$ & $0.54$ \\\cline{2-12}
    & FRR & $0.28$ & $0.64$ & $0.79$ & $0.00$ & $0.27$ & $0.00$ & $0.69$ & $0.32$ & $0.54$ & $0.43$ \\\cline{2-12}
    & HTER & \multicolumn{2}{c|}{$0.46$} & \multicolumn{2}{c|}{$0.40$} & \multicolumn{2}{c|}{$0.14$} & \multicolumn{2}{c|}{$0.51$} & \multicolumn{2}{c||}{$0.49$} \\\hline\hline
    \multirow{8}{*}{\cellcolor{gray!25}{\rotatebox[origin=c]{90}{\textbf{Replay Attack}}}} & \multirow{2}{*}{\textbf{Metric}} & \multicolumn{2}{c|}{\cellcolor{blue!25}{\textbf{MSSpoof}}} & \multicolumn{2}{c|}{\cellcolor{green!25}{\textbf{3DMAD}}} & \multicolumn{2}{c|}{\cellcolor{yellow!25}{\textbf{CSMAD}}} & \multicolumn{2}{c|}{\cellcolor{red!25}{\textbf{Our Dataset}}} & \multicolumn{2}{c||}{\cellcolor{gray!25}{\textbf{Replay Attack}}} \\ \cline{3-12}
    &  & \textit{B} & \textit{A} & \textit{B} & \textit{A} & \textit{B} & \textit{A} & \textit{B} & \textit{A} & \textit{B} & \textit{A} \\ \cline{2-12}
    & Precision & $0.71$ & $0.51$ & $0.59$ & $1.00$ & $0.70$ & $1.00$ & $0.70$ & $0.62$ & $0.97$ & $0.93$ \\\cline{2-12}
    & Recall & $0.06$ & $0.98$ & $1.00$ & $0.31$ & $1.00$ & $0.58$ & $0.52$ & $0.77$ & $0.93$ & $0.98$ \\\cline{2-12}
    & $F_1$ Score & $0.11$ & $0.67$ & $0.74$ & $0.47$ & $0.82$ & $0.73$ & $0.60$ & $0.69$ & $0.95$ & $0.95$ \\\cline{2-12}
    & FAR & $0.94$ & $0.02$ & $0.00$ & $0.69$ & $0.00$ & $0.42$ & $0.48$ & $0.23$ & $0.07$ & $0.02$ \\\cline{2-12}
    & FRR & $0.02$ & $0.94$ & $0.69$ & $0.00$ & $0.42$ & $0.00$ & $0.23$ & $0.48$ & $0.02$ & $0.07$ \\\cline{2-12}
    & HTER & \multicolumn{2}{c|}{$0.480$} & \multicolumn{2}{c|}{$0.345$} & \multicolumn{2}{c|}{$0.210$} & \multicolumn{2}{c|}{$0.355$} & \multicolumn{2}{c||}{$0.045$} \\\hline\hline
    \multirow{8}{*}{\cellcolor{red!25}{\rotatebox[origin=c]{90}{\textbf{Our Dataset}}}} & \multirow{2}{*}{\textbf{Metric}} & \multicolumn{2}{c|}{\cellcolor{blue!25}{\textbf{MSSpoof}}} & \multicolumn{2}{c|}{\cellcolor{green!25}{\textbf{3DMAD}}} & \multicolumn{2}{c|}{\cellcolor{yellow!25}{\textbf{CSMAD}}} & \multicolumn{2}{c|}{\cellcolor{red!25}{\textbf{Our Dataset}}} & \multicolumn{2}{c||}{\cellcolor{gray!25}{\textbf{Replay Attack}}} \\ \cline{3-12}
    &  & \textit{B} & \textit{A} & \textit{B} & \textit{A} & \textit{B} & \textit{A} & \textit{B} & \textit{A} & \textit{B} & \textit{A} \\ \cline{2-12}
    & Precision & $0.80$ & $0.53$ & $0.51$ & $1.00$ & $0.70$ & $1.00$ & $0.80$ & $0.89$ & $0.62$ & $0.95$ \\\cline{2-12}
    & Recall & $0.13$ & $0.97$ & $1.00$ & $0.03$ & $1.00$ & $0.58$ & $0.90$ & $0.77$ & $0.98$ & $0.41$ \\\cline{2-12}
    & $F_1$ Score & $0.22$ & $0.69$ & $0.68$ & $0.06$ & $0.82$ & $0.73$ & $0.85$ & $0.83$ & $0.76$ & $0.57$ \\\cline{2-12}
    & FAR & $0.87$ & $0.03$ & $0.00$ & $0.97$ & $0.00$ & $0.42$ & $0.10$ & $0.23$ & $0.02$ & $0.59$ \\\cline{2-12}
    & FRR & $0.03$ & $0.87$ & $0.97$ & $0.00$ & $0.42$ & $0.00$ & $0.23$ & $0.10$ & $0.59$ & $0.02$ \\\cline{2-12}
    & HTER & \multicolumn{2}{c|}{$0.450$} & \multicolumn{2}{c|}{$0.485$} & \multicolumn{2}{c|}{$0.210$} & \multicolumn{2}{c|}{$0.165$} & \multicolumn{2}{c||}{$0.305$} \\\hline
    
    \multirow{8}{*}{\cellcolor{orange!25}{\rotatebox[origin=c]{90}{\textbf{Fused Dataset}}}} & \multirow{2}{*}{\textbf{Metric}} & \multicolumn{2}{c|}{\cellcolor{blue!25}{\textbf{MSSpoof}}} & \multicolumn{2}{c|}{\cellcolor{green!25}{\textbf{3DMAD}}} & \multicolumn{2}{c|}{\cellcolor{yellow!25}{\textbf{CSMAD}}} & \multicolumn{2}{c|}{\cellcolor{red!25}{\textbf{Our Dataset}}} & \multicolumn{2}{c||}{\cellcolor{gray!25}{\textbf{Replay Attack}}} \\ \cline{3-12}
    &  & \textit{B} & \textit{A} & \textit{B} & \textit{A} & \textit{B} & \textit{A} & \textit{B} & \textit{A} & \textit{B} & \textit{A} \\ \cline{2-12}
    & Precision & $0.51$ & $0.95$ & $1.00$ & $1.00$ & $0.70$ & $1.00$ & $0.67$ & $0.72$ & $0.96$ & $1.00$ \\\cline{2-12}
    & Recall & $1.00$ & $0.03$ & $1.00$ & $1.00$ & $1.00$ & $0.58$ & $0.76$ & $0.63$ & $1.00$ & $0.96$ \\\cline{2-12}
    & $F_1$ Score & $0.68$ & $0.06$ & $1.00$ & $1.00$ & $0.82$ & $0.73$ & $0.71$ & $0.67$ & $0.98$ & $0.98$ \\\cline{2-12}
    & FAR & $0.00$ & $0.97$ & $0.00$ & $0.00$ & $0.00$ & $0.42$ & $0.24$ & $0.37$ & $0.00$ & $0.04$ \\\cline{2-12}
    & FRR & $0.97$ & $0.00$ & $0.00$ & $0.00$ & $0.42$ & $0.00$ & $0.37$ & $0.24$ & $0.04$ & $0.00$ \\\cline{2-12}
    & HTER & \multicolumn{2}{c|}{$0.490$} & \multicolumn{2}{c|}{$0.000$} & \multicolumn{2}{c|}{$0.210$} & \multicolumn{2}{c|}{$0.310$} & \multicolumn{2}{c||}{$0.020$} \\
    \Xhline{3\arrayrulewidth}
  \end{tabular}
  \caption{
    \label{table:Cross_Database}
    Performance Metrics for Training on one dataset and evaluating on another}
\end{table*}

\subsection{Test results on a Fused Dataset}

\begin{figure*}
    \centering
    \includegraphics[width=0.8\textwidth]{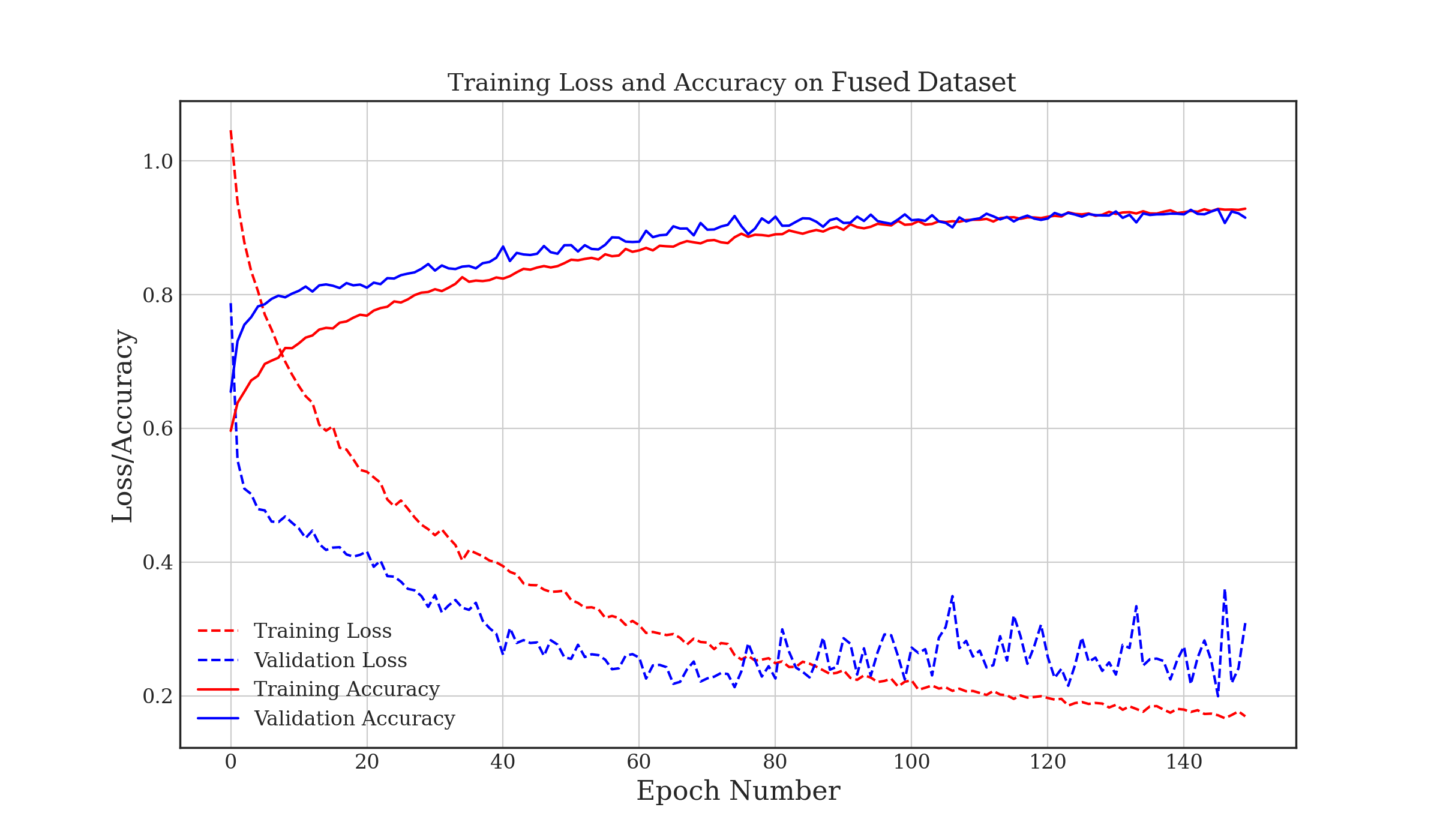}
    \caption{Loss and accuracy curves for 150 epochs while training on \textit{Fused} dataset.}
    \label{fig:fused_training}
\end{figure*}

To enhance our model's generalization capabilities, we conducted an experiment involving the training and testing the \textit{AttackNet} model on a fused dataset. This dataset combines elements from MSSpoof, 3DMAD, CSMAD, Our Dataset, and Replay Attack, providing a rich and varied pool of data reflective of numerous real-world scenarios. The results of this comprehensive testing are depicted in Table \ref{table:Cross_Database} (at the bottom) and \Cref{fig:fused_training}, offering a holistic view of the model's performance across different datasets.

Table \ref{table:Cross_Database} (at the bottom) presents the performance metrics of the model when trained on this fused dataset:

\begin{itemize}

\item \textbf{Precision and Recall:} The model demonstrated varied precision and recall rates across different datasets. For instance, on the MSSpoof dataset, it achieved a high precision of 0.95 for Attacker but a low recall of 0.03, suggesting a tendency to miss genuine cases. Conversely, on the 3DMAD dataset, precision and recall were perfect (1.0), indicating a solid ability to classify real and attacker instances correctly.

\item \textbf{$F_1$ Score:} The $F_1$ scores fluctuated across datasets, with perfect scores on 3DMAD and high scores on Replay Attack (0.98 for both Bonafide and Attacker). This variance in $F_1$ score reflects the model's changing ability to balance precision and recall under different data conditions.

\item \textbf{FAR and FRR:} Notably, the model exhibited a zero FAR and a high FRR of 0.97 on the MSSpoof dataset. This imbalance indicates a cautious approach by the model, preferring to err on the side of rejecting genuine instances rather than accepting false ones.

\item \textbf{HTER:} The Half Total Error Rate varied, with the lowest being 0.00 on the 3DMAD dataset and the highest 0.49 on the MSSpoof dataset. These figures underscore the challenges in achieving consistent error rates across diverse data sets.

\end{itemize}

\Cref{fig:fused_training} illustrates the loss and accuracy curves for training and validation over 150 epochs. The curves reveal several key observations:

\begin{itemize}

\item \textbf{Training Curve:} The training curve exhibits a steady decrease in loss and a corresponding increase in accuracy, indicating effective learning by the model. The consistent improvement over epochs suggests that the model successfully extracts and learns relevant features from the fused dataset.

\item \textbf{Validation Curve:} The validation curve, while mirroring the training curve to some extent, displays more fluctuation. This indicates the model encountering and adapting to new, unseen data in the validation set. The divergence between the training and validation curves, especially in later epochs, hints at potential overfitting issues that must be addressed in future model iterations.

\end{itemize}

Thus, the experiment with the fused dataset provides valuable insights into the model's performance in a more realistic and challenging environment. While the model shows promise in specific datasets, the variability in performance across different datasets highlights the ongoing challenge of creating a universally robust model for liveness detection. For instance, the MSSpoof dataset's distinct characteristics led to notably poor performance when included in our fused dataset testing, with a high FRR of 0.97 and low recall for Bonafide instances. This suggests a misalignment between the model's learned features and the unique aspects of MSSpoof. Excluding MSSpoof from testing yields more consistent and favorable results across other datasets, indicating its specific format may not be compatible with our model's generalization capabilities.

Future work will focus on refining the model and its training process to enhance its generalization capabilities, aiming to develop a system that performs consistently and reliably across various biometric spoofing scenarios.

\subsection{Attention Maps}

\begin{figure*}
\begin{center}
\begin{tabular}{>{\centering\arraybackslash} m{0.25cm} cc}
& \textcolor{red}{\xmark} \textbf{Attack Attention} & \textcolor{green}{\cmark} \textbf{Bonafide Attention} \\ 
 \parbox{2cm}{\vspace{-120pt} \cellcolor{blue!25}{\rotatebox[origin=c]{90}{\textbf{MSSPoof}}}} & \includegraphics[width=.28\linewidth]{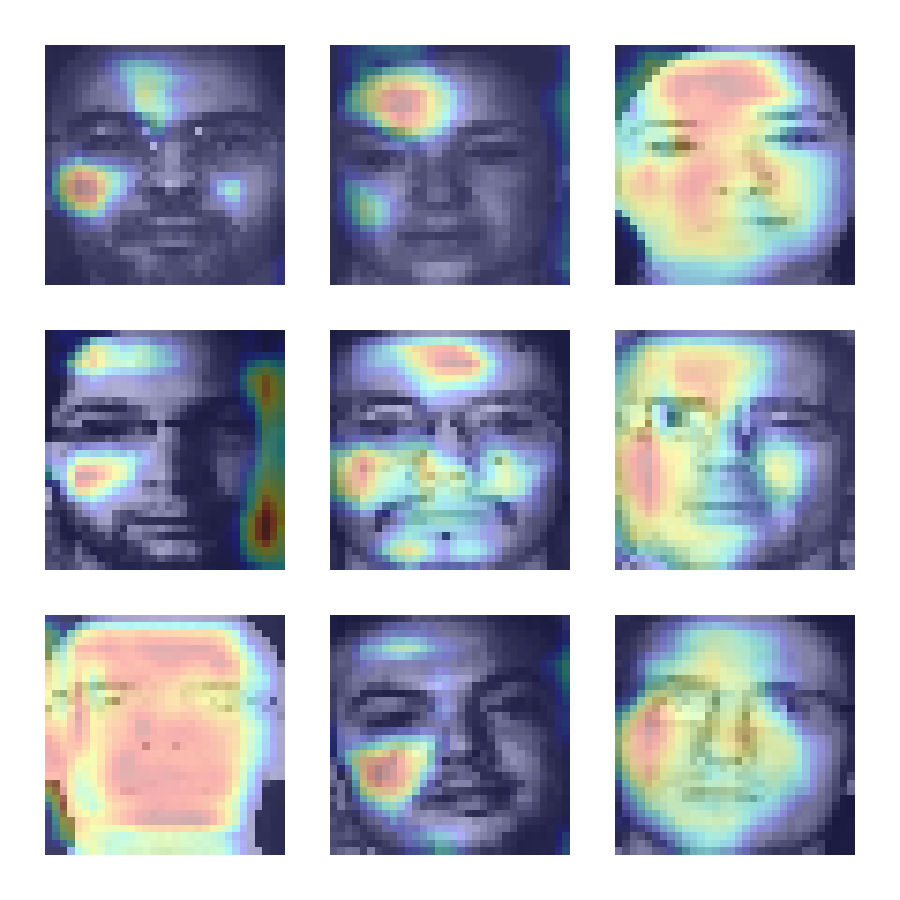} & \includegraphics[width=.28\linewidth]{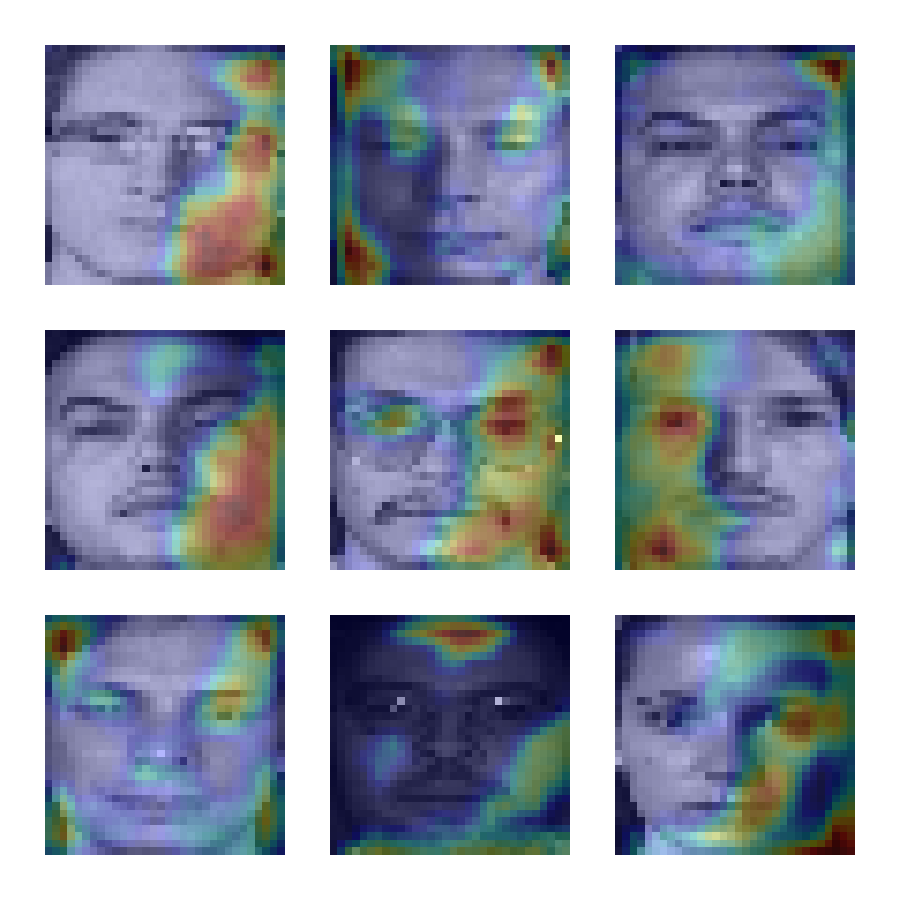} \\
 \parbox{2cm}{\vspace{-120pt} \cellcolor{green!25}{\rotatebox[origin=c]{90}{\textbf{3DMAD}}}} & \includegraphics[width=.28\linewidth]{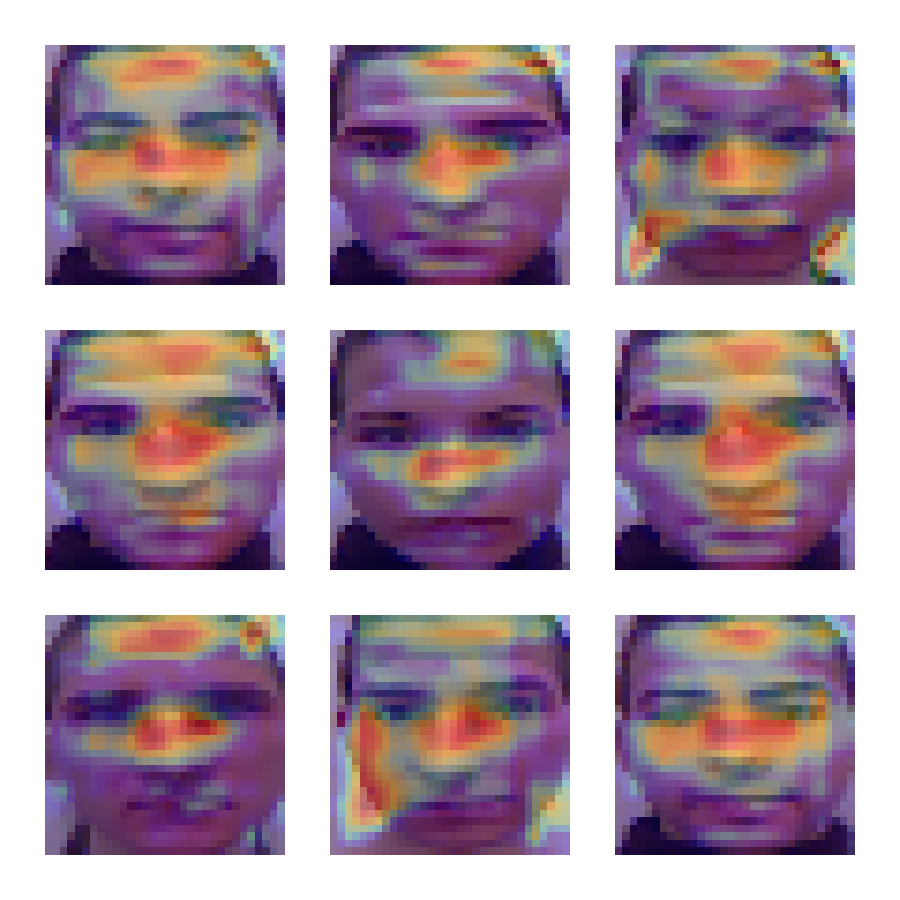} & \includegraphics[width=.28\linewidth]{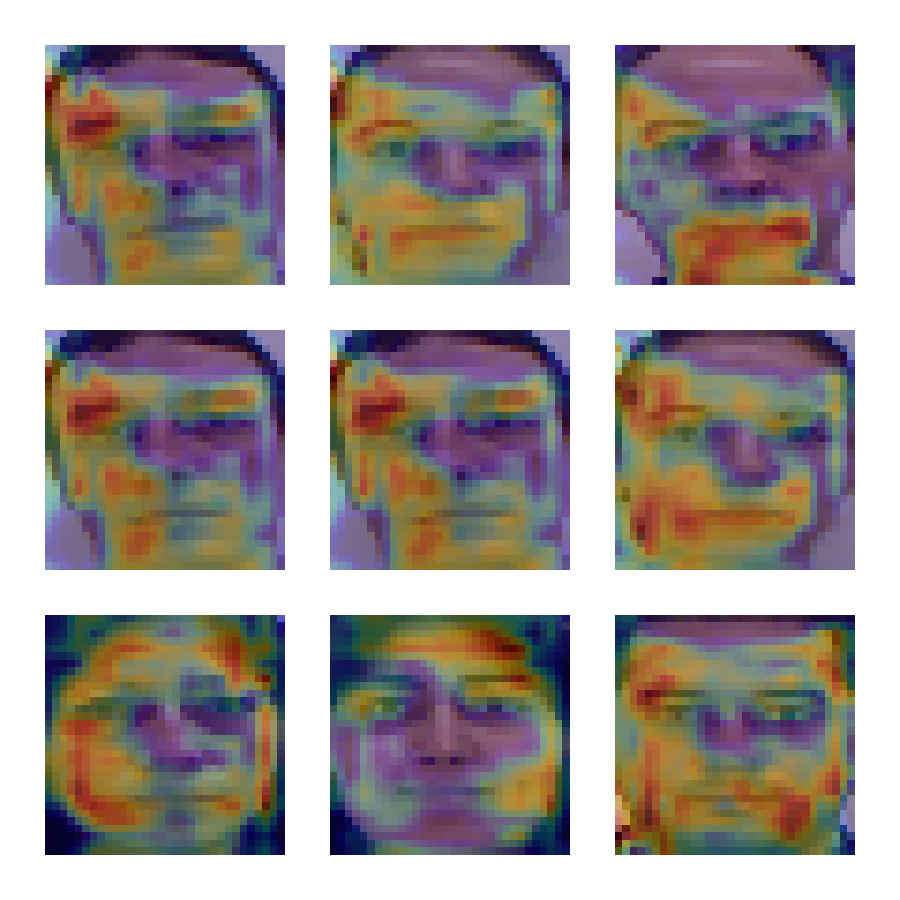} \\
 \parbox{2cm}{\vspace{-120pt} \cellcolor{gray!25}{\rotatebox[origin=c]{90}{\textbf{Replay Attack}}}} & \includegraphics[width=.28\linewidth]{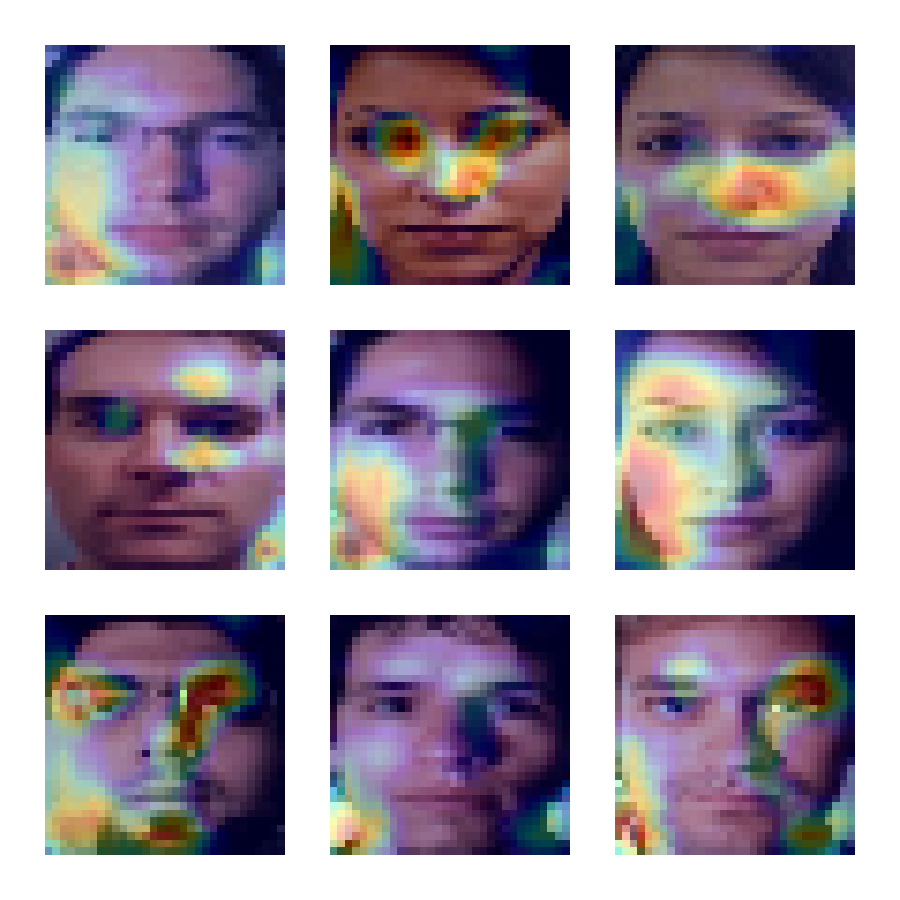} & \includegraphics[width=.28\linewidth]{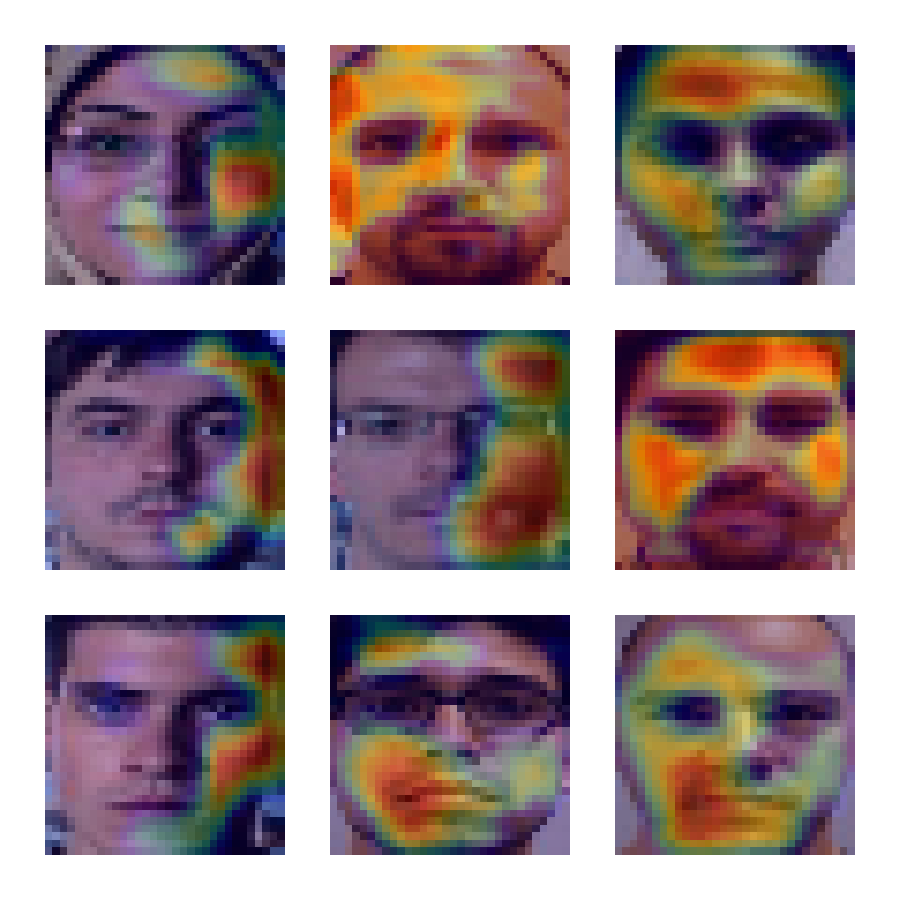} \\
 \parbox{2cm}{\vspace{-120pt} \cellcolor{red!25}{\rotatebox[origin=c]{90}{\textbf{Our Dataset}}}} & \includegraphics[width=.28\linewidth]{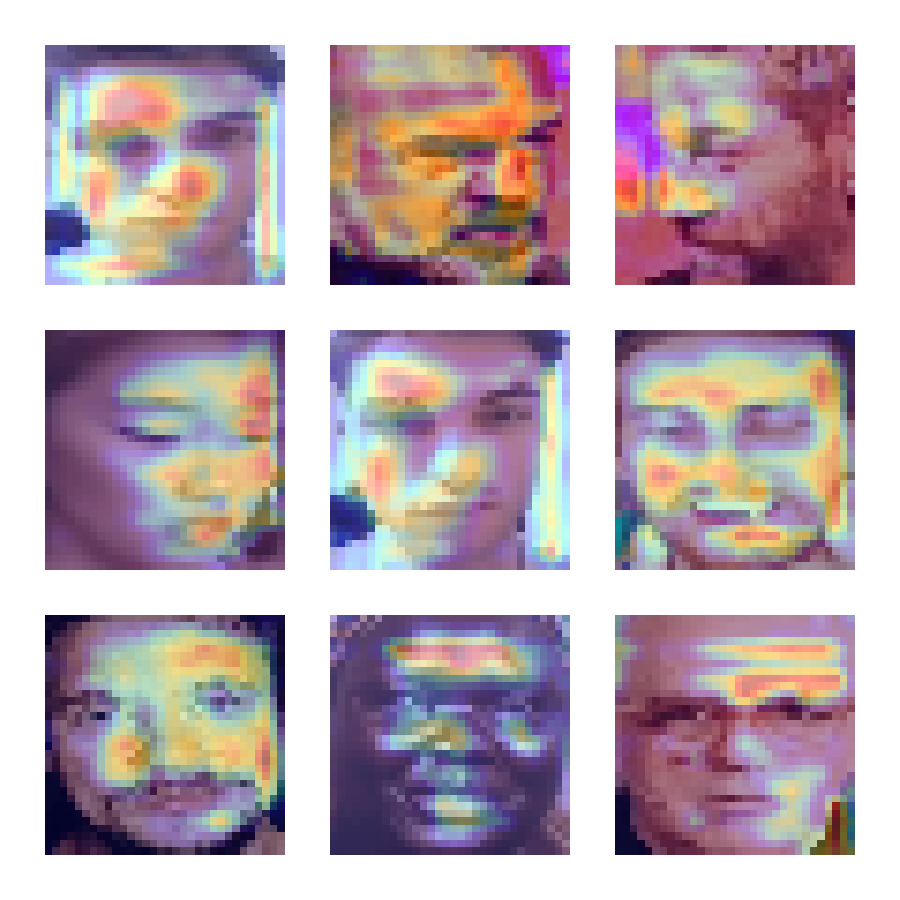} & \includegraphics[width=.28\linewidth]{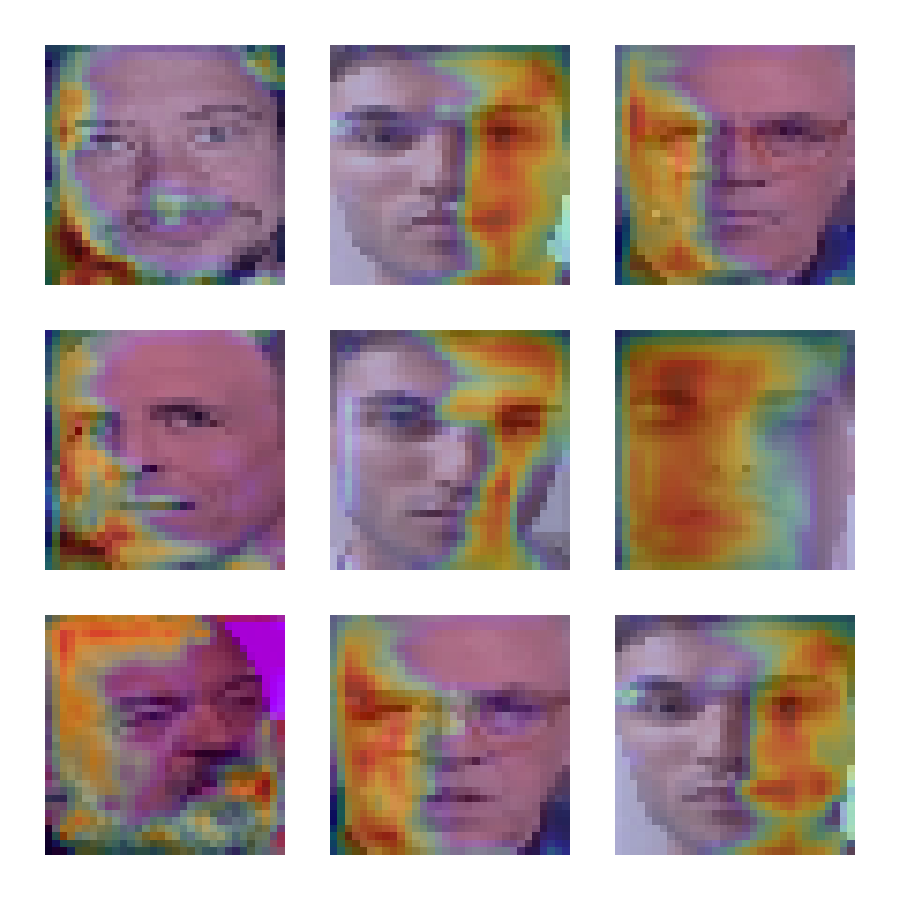}
\end{tabular}
\end{center}
\caption{Attention maps built using \textit{Grad-CAM} technology for four datasets. Highlighted regions show activations that contributed the most to the classification decision, where temperature represents the intensity. \textbf{Left} column -- what led to ``attack'' decision, \textbf{right} -- what led to ``bonafide'' decision.}
\label{fig:grad-cam}
\end{figure*}

To get a grasp of what the neural network has learned, we have used  Gradient-weighted Class Activation Mapping (\textit{Grad-CAM}) technology by \citet{selvaraju2017grad} using \texttt{tf-keras-vis} library by \citet{Kubota_tf-keras-vis_2023}. This technique visually explains what regions the neural network perceives as ``important'' by building the colormap with the same shape as the input image. After applying our trained model to four datasets, we obtained the \Cref{fig:grad-cam}. 

In most cases, the neural network seems to try to make predictions based on the skin texture, which is quite reasonable: the only difference between an authentic photo and one recording another screen or physical image is either certain lightning inconsistencies or simply the surface texture. For example, oftentimes, \textit{MSSPoof} makes decisions based on the cheek, while in the \textit{Replay Attack} the primary reason for detecting an image as fake is the lightning inconsistency in the corner. 

Probably the most exciting result is for the \textit{3DMAD} dataset. Here, the primary reason for the neural network to decide that the image is fake is based on the nose. 

As can be seen, the features which the neural network perceives to be important for classification is different for each of the dataset, which might partially explain such drastic changes when sometimes the model trained on one dataset cannot generalize well on another datasets.

\section{Discussion}\label{section:discussion}

The realm of biometric spoofing detection has witnessed an array of innovative approaches, particularly in the context of cross-database testing. This section delves into the methodologies employed in various significant studies and juxtaposes their results with ours, offering a comprehensive perspective on the state-of-the-art in spoofing detection.

\subsection{In-Depth Analysis of Cross-Database Testing Methods}

Recent works by Arora et al. \citep{framework_spoofing}, Sun et al. \citep{fcn_detection}, Wang et al., \citep{visual_saliency}, and others have explored various detection techniques:

\begin{itemize}

\item \textbf{FC–CNN:} Utilizes the last fully connected layer of CNN for feature extraction. While effective in feature representation, its adaptability across different datasets can be limited.

\item \textbf{Multi-scale LBP (MS-LBP):} This method concatenates three types of LBP histograms. It is proficient in capturing texture but may not fully encapsulate more subtle spoofing cues.

\item \textbf{Image Distortion Analysis (IDA):} Focuses on analyzing specular reflection and color diversity differences. It provides a unique perspective but may overlook other crucial spoofing indicators.

\item \textbf{Videolet:} Combines multi-scale LBP with HOOF operators for texture and motion analysis. Its reliance on motion may limit its effectiveness against static spoofs.

\item \textbf{Color Texture (CT):} Extracts adjacent LBP co-occurrence in different color spaces. While innovative, its performance can vary based on the color properties of datasets.

\item \textbf{CDCN++:} Employs central difference convolution for enhanced feature extraction. This method shows promise but can be computationally intensive.

\item \textbf{VSFM:} Integrates Visual Saliency with Facial Motion characteristics. This comprehensive dual approach may not generalize well across datasets with varying motion dynamics.

\item \textbf{IQM+IDA+SVM:} A combination of Image Quality Measure, Image Distortion Analysis, and SVMs. This method provides a balanced approach but may struggle with high-quality spoofs.

\item \textbf{Auxiliary:} Uses spatial and temporal information, but its performance can be contingent on the dataset's inherent characteristics.

\item \textbf{Noise Modeling:} A novel approach that decomposes a spoof into noise and a live face. Its effectiveness depends on the model's ability to separate these components accurately.

\item \textbf{STASN:} Focuses on spatio-temporal aspects with attention mechanisms. While powerful, it requires substantial computational resources.

\item \textbf{SAPLC:} Aggregates pixel-level classifiers in a spatial context. This method is innovative but may be challenged by complex spoofing attacks.

\item \textbf{Robust Framework:} Leverages dimensionality reduction and convolutional autoencoders. It's effective yet may not capture all nuances of spoofing.

\end{itemize}

\begin{table*}
\centering
\begin{tabular}{ccccc}
\Xhline{3\arrayrulewidth}
\textbf{Publication} & \textbf{AI Model Used} & \textbf{Training Dataset} & \textbf{Testing Dataset} & \textbf{HTER} \\ \hline
\citep{yang2014}  & FC–CNN & SMFMVD & SMAD & 29\% \\ \hline
\citep{yang2014}  & FC–CNN & SMAD & SMFMVD & 31\% \\ \hline
\citep{erdogmus2014}  & MS-LBP & SMFMVD & SMAD & 53\% \\ \hline
\citep{erdogmus2014} & MS-LBP & SMAD & SMFMVD & 44\% \\ \hline
\citep{wen2015} & IDA & SMFMVD & SMAD & 37\% \\ \hline
\citep{wen2015} & IDA & SMAD & SMFMVD & 47\% \\ \hline
\citep{siddiqui2016}  & Videolet & SMFMVD & SMAD & 29\% \\ \hline
\citep{siddiqui2016}  & Videolet & SMAD & SMFMVD & 44\% \\ \hline
\citep{boulkenafet2016}  & CT & SMFMVD & SMAD & 30\% \\ \hline
\citep{boulkenafet2016}  & CT & SMAD & SMFMVD & 49\% \\ \hline
\citep{yu2020}  & CDCN++ & SMFMVD & SMAD & 49\% \\ \hline
\citep{yu2020}  & CDCN++ & SMAD & SMFMVD & 46\% \\ \hline
\citep{visual_saliency} & VSFM & SMFMVD & SMAD & 22\% \\ \hline
\citep{visual_saliency} & VSFM & SMAD & SMFMVD & 20\% \\ \hline
\citep{nikisins2018}  & IQM+IDA+SVM & CASIA-FASD & Replay-Attack & 37\% \\ \hline
\citep{nikisins2018} & IQM+IDA+SVM & Replay-Attack & CASIA-FASD & 41\% \\ \hline
\citep{liu2018} & Auxiliary & CASIA-FASD & Replay-Attack & 28\% \\ \hline
\citep{liu2018} & Auxiliary & Replay-Attack & CASIA-FASD & 28\% \\ \hline
\citep{jourabloo2018}  & Noise Modeling & CASIA-FASD & Replay-Attack & 29\% \\ \hline
\citep{jourabloo2018}  & Noise Modeling & Replay-Attack & CASIA-FASD & 41\% \\ \hline
\citep{yang2019}  & STASN & CASIA-FASD & Replay-Attack & 32\% \\ \hline
\citep{yang2019} & STASN & Replay-Attack & CASIA-FASD & 31\% \\ \hline
\citep{fcn_detection} & SAPLC & CASIA-FASD & Replay-Attack & 27\% \\ \hline
\citep{fcn_detection} & SAPLC & Replay-Attack & CASIA-FASD & 38\% \\ \hline
\citep{framework_spoofing} & Robust Framework & 3DMAD & CASIA & 40\% \\ \hline
\citep{framework_spoofing} & Robust Framework & CASIA & Replay Attack & 41\% \\ \hline

Our work & \textit{AttackNet} & MSSpoof & 3DMAD & \textbf{31\%} \\ \hline
Our work & \textit{AttackNet} & MSSpoof & CSMAD & \textbf{26.5\%} \\ \hline
Our work & \textit{AttackNet} & 3DMAD  & CSMAD & \textbf{13.5\%} \\ \hline
Our work & \textit{AttackNet} & Replay Attack  & CSMAD & \textbf{21\%} \\ \hline
Our work & \textit{AttackNet} & Our Dataset  & Replay Attack & \textbf{31\%} \\ \hline
Our work & \textit{AttackNet} & Our Dataset  & CSMAD & \textbf{21\%} \\ \Xhline{3\arrayrulewidth}

\end{tabular}
\caption{Comparative Analysis of Cross-Database Testing in Spoofing Detection}
\end{table*}

The comparative analysis reveals a diverse range of HTER values across studies, highlighting the varying effectiveness of different models and techniques in cross-database scenarios. While some models demonstrate low HTER, indicating high adaptability and robustness, others exhibit higher HTER values, underscoring challenges in generalizing across datasets. Our study contributes to this landscape by offering a nuanced understanding of how a model trained on a combined dataset can perform across multiple datasets, reflecting a balance between specificity and adaptability.

\subsection{Comparing Our Results with Established Studies}

The comparative analysis of our cross-database testing results with those from established studies in the field reveals significant insights into the performance of various AI models in spoofing detection. Our study aimed to understand the robustness and adaptability of models across different datasets, a crucial aspect in developing effective anti-spoofing solutions.

\subsubsection{Analysis of Cross-Database Testing in Prior Research}

Publications like \citep{yang2014} using FC-CNN and \citep{face_spoofing_attacks} employing MS-LBP showed varying levels of HTER across different datasets. For instance, Yang et al. \citep{yang2014} reported HTERs of 31\% and 32\% when training and testing across SMFMVD and SMAD datasets, respectively. This variability is indicative of the challenges faced in model generalization across different datasets.

More recent approaches like CDCN++ by Yu et al. \citep{yu2020} and VSFM by Wang et al. \citep{visual_saliency} demonstrated improved but still varied HTERs, suggesting advancements in model robustness yet highlighting persisting challenges in cross-database adaptability.

\subsubsection{Our Cross-Database Testing Outcomes}

Our study observed similar trends, where models trained on one dataset, such as MSSpoof, and tested on others, like 3DMAD or CSMAD, exhibited fluctuating performance metrics. For example, training on the MSSpoof dataset resulted in an HTER of 0.06 on the same dataset but increased to 0.305 and 0.415 when tested on 3DMAD and Our Dataset, respectively.
These findings align with the trends observed in the established studies, emphasizing the ongoing challenge of achieving consistent model performance across varied datasets.

\subsubsection{Performance on Combined Dataset}

\begin{figure}
    \centering
    \includegraphics[width=0.5\textwidth]{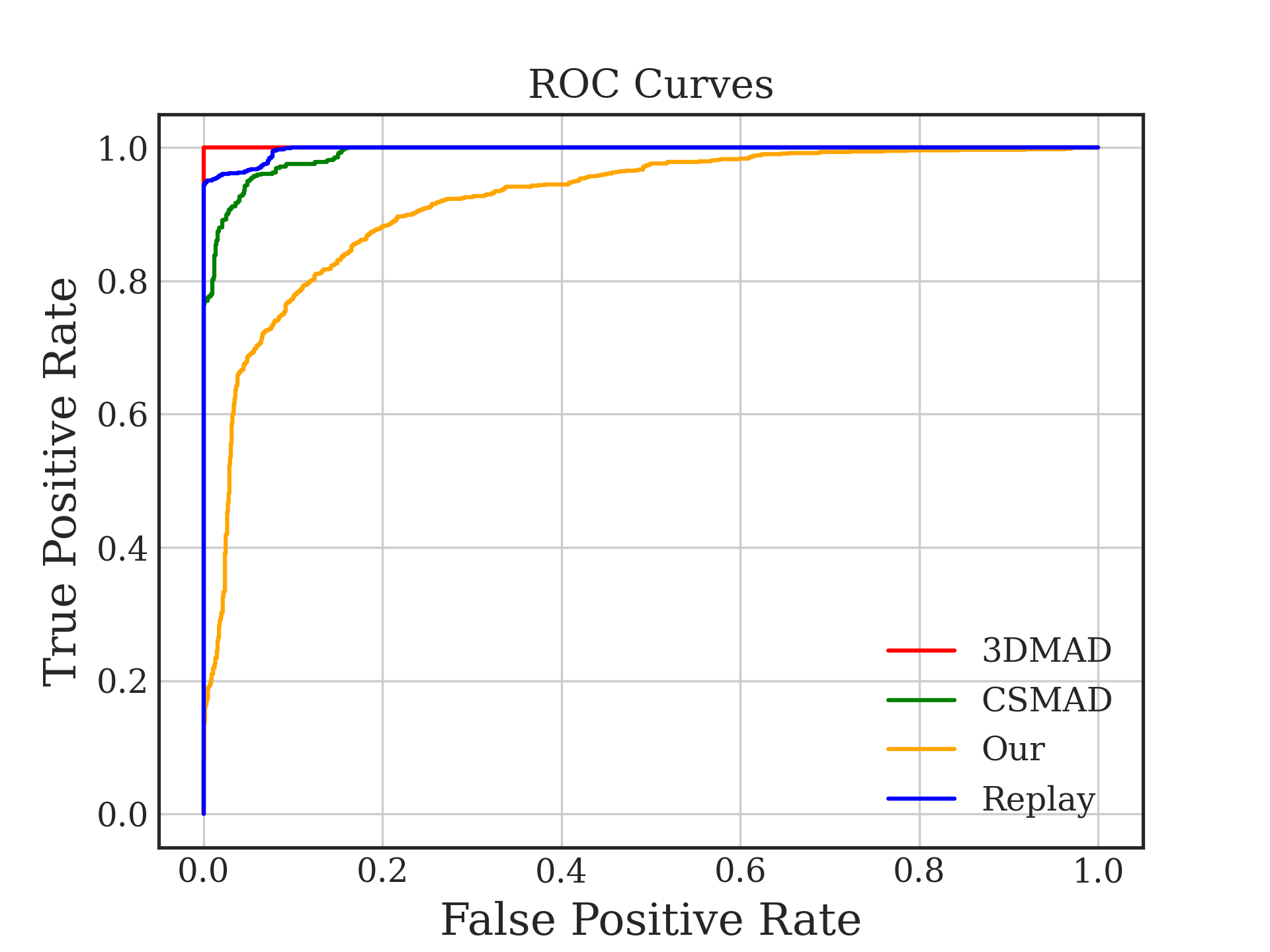}
    \caption{ROC curves of the model trained on the fuzed dataset and then tested on datasets used. Unfortunately, the model did not generalize well on the MSSPoof dataset, so it is not included in the figure.}
    \label{fig:roc}
\end{figure}

In contrast, when our model was trained on a combined dataset, incorporating elements from MSSpoof, 3DMAD, CSMAD, Our Dataset, and Replay Attack, the performance metrics showed a remarkable improvement. For instance, the HTER on the Our Dataset reduced to 0.31, demonstrating enhanced adaptability and robustness. When tested on the 3DMAD and Replay Attack datasets, there is almost zero error rate, although the HTER is still high for the MSSpoof dataset. Finally, we plotted ROC curves, displayed in \Cref{fig:roc}, which visualize the resultant model's performance.

This improvement in our model's performance when trained on a combined dataset resonates with the need for diverse training, as observed in other studies. It highlights the potential benefits of exposing the model to a wide range of data characteristics during training to improve its generalization capabilities.

\subsubsection{Scalability to Millions of Users: Predictive Analysis and System Robustness} 

A critical aspect of deploying any biometric authentication system, such as the one proposed in our study, is its scalability. Our system's design inherently supports scalability, thanks to its lightweight architecture and the efficient processing capabilities demonstrated in the earlier sections. Specifically, the model's low computational overhead, quantified through minimal Floating Point Operations (FLOPs) and a compact parameter set, signifies its potential to operate within large-scale systems without a proportional increase in computational resources.

The key to successful scaling also lies in the adaptive learning capabilities of our model. Implementing continuous learning mechanisms, where the model periodically updates itself with new data, could mitigate the risk of obsolescence due to evolving attack methodologies. This approach ensures that the system remains effective not only across a vast number of users but also over time.

Lastly, addressing scalability involves considering the infrastructure supporting the deployment of our model. Cloud-based architectures and distributed computing resources can dynamically adjust to the demand, ensuring that the increase in user numbers correlates with available processing power. This flexibility, combined with our model's efficiency, forms a solid foundation for scaling up to millions of users.

In summary, while direct empirical evidence of our system's performance at the scale of millions of users remains beyond the scope of this paper, predictive analysis and the inherent design of our model provide a strong basis for its potential scalability. Future work will focus on empirical validation through pilot deployments in large-scale environments, further refining the model and infrastructure to ensure reliability and effectiveness at any scale.

\section{Conclusion}\label{section:conclusion}

Our comprehensive research journey in liveness detection has led us to significant insights and advancements. Central to our findings is the development and evaluation of \textit{AttackNet}, a model characterized by its minimal parameter footprint and substantial performance, making it highly suitable for mobile and embedded devices deployment.

\subsection{Model Performance and Benchmarking}

As detailed in the performance analysis section and illustrated in Table \ref{table:performance}, \textit{AttackNet} demonstrates an impressive efficiency with only 22.7 MFLOPs and a mere 0.3M parameters. This lean architecture favors it against more parameter-intensive models like \textit{MobileNetV3} and \textit{ShuffleNet}. The comparative analysis underscores \textit{AttackNet}'s potential for applications where computational resources are limited, offering an optimal balance between efficiency and effectiveness.

\subsection{Overcoming the Challenge of Cross-Database Testing}

A pivotal aspect of our study was addressing the challenge of cross-database testing. Traditional models, while performing adequately on their training datasets, often faltered when exposed to unfamiliar data. Through our rigorous testing regime, including training on a combined dataset, we made significant strides in enhancing the model's adaptability and robustness across various datasets. This approach marked a notable improvement in the model's generalization capabilities, as evidenced by the reduced error rates and improved performance metrics across different testing scenarios.

\subsection{Future Directions and Research Avenues}

Our research is a foundation for various promising avenues in advancing anti-spoofing and liveness detection technologies. Moving forward, we envision several key areas of focus that could significantly enhance the efficacy and applicability of our model:

\begin{itemize}

\item \textbf{Advanced Architectural Innovations:} Building on the solid base of \textit{AttackNet}, we aim to explore integrating state-of-the-art techniques such as attention mechanisms and few-shot learning. These sophisticated approaches can potentially augment the model's capability to distinguish between genuine and spoofed biometric data more effectively, enhancing its adaptability to various spoofing scenarios.

\item \textbf{Expanded Dataset Integration:} A critical aspect of our future work is the continuous expansion and diversification of the training datasets. By incorporating a more comprehensive array of data, particularly from more challenging and varied sources, we can further push the boundaries of our model’s robustness and reliability. This expansion will also facilitate the model's ability to adapt to the evolving nature of spoofing attacks.

\item \textbf{Real-World Application Testing:} Deploying \textit{AttackNet} in practical, real-world scenarios is a vital step. This real-time testing will offer invaluable insights into the model's operational effectiveness and provide opportunities for iterative refinements based on real-world feedback and performance metrics.

\item \textbf{Efficiency Optimization for Embedded Systems:} While \textit{AttackNet} already exhibits significant efficiency, optimizing it for deployment on resource-constrained devices remains a priority. Future iterations will focus on reducing computational demands while maintaining high accuracy, ensuring suitability for integration into a wide range of embedded systems.

\item \textbf{Image Fusion Techniques:} This approach can enhance the model's performance by combining multiple images or features into a single, more informative representation. This method could prove especially beneficial in scenarios where a single image might not provide sufficient data for accurate spoof detection.

\item \textbf{Exploration of Additional Techniques:} We are also considering the application of other advanced techniques, such as adversarial training, to fortify the model against sophisticated spoofing methods further. These enhancements address the dynamic and increasingly complex nature of spoofing attacks.
\item \textbf{Looking into quality impact:} An exciting question is how the quality of a photo affects how accurately it is classified via spoofing detection frameworks. While some studies have examined how well face recognition works with different image qualities, like \citep{face-recognition-quality}, few have checked how well spoofing detection works across various databases (for example, for five datasets from \Cref{section:methodology}) when the image quality changes. Besides, having the datasets addressing these issues would significantly boost performance evaluation metrics.
\item \textbf{Testing AttackNet as the part of multibiometrics system}: For enhancing security, multiple face photos or biometric samples might be provided as an input, as noted in \citep{almahafzah2012survey}. Despite the existence of CNN approaches like \citep{talreja2017multibiometric}, to our knowledge, there are still no deep-learning-based frameworks (at least, open-source ones) capable of combining different samples effectively.
\end{itemize}

In summary, the progression of our research will focus on both technological innovation and practical application. By continuously refining \textit{AttackNet} and adapting it to the ever-changing landscape of biometric security threats, we aspire to contribute significantly to developing more secure, efficient, and reliable biometric authentication systems. Our commitment to this research reflects our understanding that the journey toward perfecting spoofing detection is an ongoing endeavor, and our work represents a crucial step in this critical field.
\section{Declarations}

\subsection*{Author contributions}

\begin{itemize}
    \item Oleksandr Kuznetsov: Conceptualization, Methodology.
    \item Dmytro Zakharov: Software and Investigation, Visualization.
    \item Emanuele Frontoni: Supervision, Review and Editing.
    \item Andrea Maranesi: Data Curation, Software and Validation.
\end{itemize}

\subsection*{Data availability}
The datasets generated during and/or analyzed during the current study are available from the corresponding author upon reasonable request.

\subsection*{Declaration of interests}
\begin{itemize}
    \item I declare that the authors have no competing financial interests or other interests that might be perceived to influence the results and/or discussion reported in this paper.
    \item The results/data/figures in this manuscript have not been published elsewhere, nor are they under consideration (from you or one of your Contributing Authors) by another publisher.
    \item All of the material is owned by the authors, and/or no permissions are required.
\end{itemize}

\subsection*{Compliance with ethical standards}
The mentioned authors have no conflict of interest in this article. This article contains no studies with human participants or animals performed by any of the authors.

\subsection*{Funding}

\begin{enumerate}
    \item This project has received funding from the \textit{European Union’s Horizon 2020} research and innovation programme under the \textit{Marie Skłodowska-Curie grant} agreement No. 101007820 - \textit{TRUST}.This publication reflects only the author’s view and the \textit{REA} is not responsible for any use that may be made of the information it contains.
    \item This research was funded by the \textit{European Union – NextGenerationEU} under the  \textit{Italian Ministry of University and Research} (MIUR), \textit{National Innovation Ecosystem grant} ECS00000041-VITALITY-CUP D83C22000710005

\end{enumerate}

\bibliographystyle{elsarticle-harv} 
\bibliography{refs}

\end{document}